\documentclass[12pt]{article}

\usepackage{graphicx}
\usepackage{xcolor}
\usepackage{subcaption}
\usepackage{amsmath,amssymb}
\usepackage{longtable}
\usepackage[letterpaper,margin=1in]{geometry}
\usepackage{array}
\usepackage[hidelinks]{hyperref}
\newcommand{\chssvideourl}{https://www.dropbox.com/scl/fi/du3qi7tvdjitqol7l4ckk/open_h_c-h-s-s_sample_12k_v1.mp4?rlkey=q5aooz898330xy1v5b1qtng2c\&st=i8jatft7\&dl=0}
\newcommand{\groothendtoendurl}{https://drive.google.com/file/d/1gXX0imqTKcQ_GOcKzq5yH31gxVTdSlHy/view?usp=sharing}
\newcommand{\groothwoundclosureurl}{https://drive.google.com/file/d/1u1jHkaSH8QZWuyZHe5NplFUGU5o4fdq6/view?usp=sharing}
\renewenvironment{abstract}
	{\quotation}
	{\endquotation}

\date{}

\makeatletter
\renewcommand{\fnum@figure}{\textbf{Figure \thefigure}}
\renewcommand{\fnum@table}{\textbf{Table \thetable}}
\makeatother

\usepackage{scicite}
\usepackage{url}

\usepackage[T1]{fontenc}
\usepackage[utf8]{inputenc}
\usepackage{lmodern}

\def\scititle{
	Open-H-Embodiment: A Large-Scale Dataset for Enabling Foundation Models in Medical Robotics
}
\title{\bfseries \boldmath \scititle}

\author{
	\parbox{\textwidth}{\centering\fontsize{7}{7}\selectfont
	Nigel Nelson$^{1, \dagger}$,
	Juo-Tung Chen$^{2, \dagger}$,
    Jesse Haworth$^{2, \dagger}$,
	Xinhao Chen$^{2, \dagger}$,
	Lukas Zbinden$^{1, \dagger}$,
	Dianye Huang$^{3, \dagger}$,
    Alaa Eldin Abdelaal$^{4}$,
    Alberto Arezzo$^{5}$,
    Ayberk Acar$^{6}$,
	Farshid Alambeigi$^{7}$
% ...
}
\\[1ex]
\\[0.5ex]
	\parbox{\textwidth}{\centering\tiny
	$^{1}$NVIDIA,
	$^{2}$Johns Hopkins University,
	$^{3}$Technical University of Munich,
	$^{4}$Stanford University,
	$^{5}$University of Turin,
	$^{6}$Vanderbilt University,
	$^{7}$The University of Texas at Austin,
% ...
    {\itshape \dag~Co-first authors. \ddag~Co-senior authors.}
}}

\begin{document}
\maketitle

\begin{abstract} \bfseries \boldmath
Autonomous medical robots hold promise to improve patient outcomes, reduce provider workload, democratize access to care, and enable superhuman precision.
However, autonomous medical robotics has been limited by a fundamental data problem: existing medical robotic datasets are small, single-embodiment, and rarely shared openly, restricting the development of foundation models that the field needs to advance. We introduce \textbf{Open-H-Embodiment}, the largest open dataset of medical robotic video with synchronized kinematics to date, spanning more than 50 institutions and multiple robotic platforms including the CMR Versius, Intuitive Surgical's da Vinci, da Vinci Research Kit (dVRK), Rob Surgical BiTrack, Virtual Incision's MIRA, Moon Surgical Maestro, and a variety of custom systems, spanning surgical manipulation, robotic ultrasound, and endoscopy procedures. We demonstrate the research enabled by this dataset through two foundation models. \textbf{GR00T-H} is the first open foundation vision-language-action model for medical robotics, which is the only evaluated model to achieve full end-to-end task completion on a structured suturing benchmark (25\% of trials vs.\ 0\% for all others) and achieves 64\% average success across a 29-step ex vivo suturing sequence. We also train \textbf{Cosmos-H-Surgical-Simulator}, the first action-conditioned world model to enable multi-embodiment surgical simulation from a single checkpoint, spanning nine robotic platforms and supporting in silico policy evaluation and synthetic data generation for the medical domain. These results suggest that open, large-scale medical robot data collection can serve as critical infrastructure for the research community, enabling advances in robot learning, world modeling, and beyond.
\end{abstract}

\definecolor{projectblue}{RGB}{15,20,40}
\vspace{0.5em}
\begin{center}
\small\color{projectblue}%
\raisebox{-0.15em}{\includegraphics[height=0.9em]{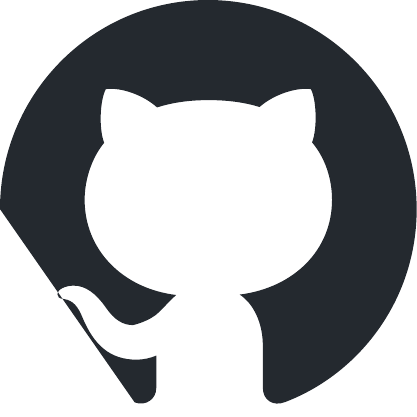}}\,%
\textbf{Project page:} \url{https://open-h.github.io/open-h-embodiment/}
\end{center}
\vspace{0.5em}

\begin{figure}
  \centering
  \includegraphics[width=0.99\textwidth]{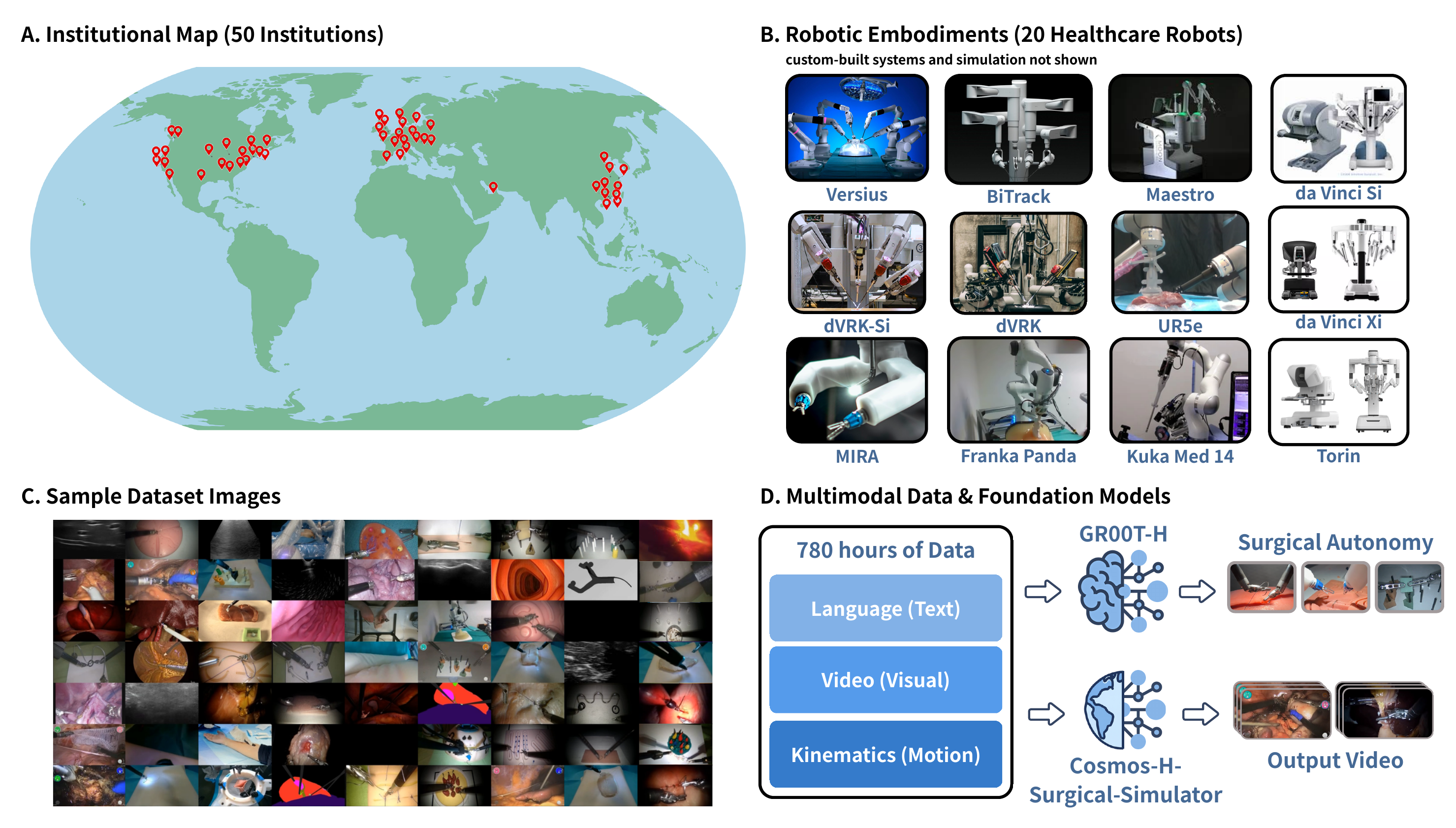}
  \caption{\textbf{Open-H-Embodiment overview.}  (\textbf{A}) Geographic distribution of the 50 participating institutions across North America, Europe, the Middle East, and Asia. (\textbf{B}) The 20 healthcare robotic platforms represented in the dataset, spanning surgical systems (da Vinci Si, da Vinci Xi, dVRK, dVRK-Si, MIRA, Versius, BiTrack, Maestro, Torin), general-purpose manipulators adapted for clinical use (Franka Panda, UR5e, Kuka Med 14), and emerging platforms. (\textbf{C}) Representative frames from the dataset illustrating the diversity of tasks, viewpoints, and tissue types covered, including robotic surgery, robotic ultrasound, and related healthcare manipulation tasks. (\textbf{D}) The dataset comprises 780 hours of synchronized multimodal demonstrations spanning language annotations, video observations, and kinematic trajectories. This corpus supports two downstream directions: training GR00T-H, a healthcare-focused vision-language-action model targeting surgical autonomy, and training Cosmos-H-Surgical-Simulator, a multi-embodiment, action-conditioned world model for surgical scene synthesis.}
  \label{fig:open_h_overview}
\end{figure}

\section*{Summary}
Open-H: The largest medical robotic dataset of paired video and kinematics enabling multi-embodiment foundation models.

\section*{Introduction}
In the global healthcare system, surgical patient outcomes vary tremendously and are heavily dependent on factors such as surgeon experience, workload and fatigue, and disparities in institutional resources.
This variability has the potential to be exacerbated by a mounting global healthcare workforce crisis; in the United States alone, a deficit of 13,000 to 86,000 surgeons is projected by 2036 \cite{aamcglobaldata2024, moffatt2018providing}.
Autonomous robotic systems offer a critical pathway to mitigate surgeon workload, reduce burnout, and expand access to high-quality care, particularly in underserved regions or ambulatory surgery centers lacking specialized expertise.
The path toward such autonomy is already underway; medical robots are increasingly deployed across the healthcare spectrum, taking on diverse roles ranging from patient rehabilitation to precise diagnostic imaging.
While many of these systems operate with significant independence, surgical robots present a stark contrast.
They remain almost entirely dependent on real-time human teleoperation, functioning at Level 1 (robot assistance) on surgical levels of autonomy~\cite{haidegger2019autonomy, schmidgall2025will}.
Developing autonomous and semi-autonomous capabilities for high-volume procedures, such as cholecystectomy and suturing, provides a vital strategy to reduce performance variability and extend surgical capacity to underserved populations \cite{kazanzides2014open}.

Recent advances in robot learning have enabled the autonomous execution of complex surgical tasks, including portions of ex vivo cholecystectomy \cite{kim2024srt} and in vivo tissue retraction and gauze packing \cite{long2025surgical}. While these results demonstrate the potential for autonomous systems in clinical procedures, current models remain fragile, failing when confronted with procedural variations and struggling to generalize across different surgical tasks. To achieve scalable autonomy, the surgical robotics field must adopt the paradigms currently driving general artificial intelligence. The broader machine learning community has established a powerful empirical principle: large-scale, diverse pretraining datasets unlock generalist models that significantly outperform narrow policies, fine-tune efficiently, and generalize to novel domains. This trend is well-established in natural language processing \cite{brown2020gpt3, devlin2019bert} and computer vision \cite{dosovitskiy2021vit, radford2021clip, he2022mae}, and is increasingly evident in general-purpose robotics \cite{openxembodiment2024}.
The evolution of Vision-Language-Action (VLA) models perfectly illustrates this trajectory. Foundational work like RT-1~\cite{brohan2023rt1} and RT-2~\cite{brohan2023rt2} demonstrated that transformer-based visuomotor policies exhibit clear scaling behavior, effectively transferring web-scale semantic knowledge to robotic control. Subsequent efforts proved the immense value of cross-embodiment data sharing. Open-X-Embodiment~\cite{openxembodiment2024} pooled over one million trajectories, enabling the RT-2-X model to improve the success rate upon the single-embodiment RT-2 by approximately 50\% on emergent tasks. This breakthrough led to a rapid progression of highly efficient, generalist architectures. Models such as Octo~\cite{octo2024}, CrossFormer~\cite{doshi2024crossformer}, and the 7B-parameter OpenVLA~\cite{kim2024openvla} showed that policies pretrained on massive corpora can be efficiently fine-tuned to new embodiments, and outperform narrowly trained models.
Frameworks like AgiBot World~\cite{agibotworld2025}, Google's Gemini Robotics~\cite{team2025gemini}, and NVIDIA's GR00T~\cite{nvidia2025groot} pushed these boundaries further, training on vast multimodal and synthetic datasets that span tens of thousands of hours without hitting performance saturation. Most recently, works such as Cosmos-Policy~\cite{kim2026cosmos}, DreamZero~\cite{ye2026worldactionmodelszeroshot}, and LingBot-VA~\cite{li2026lingbotva} have shifted from semantic to spatiotemporal priors, replacing language-pretrained backbones with video foundation models that internalize physical dynamics from internet-scale video and show strong initial results on general manipulation benchmarks. The central lesson across this progression is that data diversity, multimodality richness, and scale compound. Each generation of larger, more varied corpora yields stronger generalization, higher absolute performance, and more efficient fine-tuning.

Surgical robotics, however, has not shared in these advances. The SutureBot benchmark~\cite{haworth2025suturebot} evaluated state-of-the-art general-purpose VLAs, including OpenVLA~\cite{kim2024openvla}, GR00T-N1~\cite{nvidia2025groot}, and $\pi_0$~\cite{black2024pi0}, on autonomous suturing and found that all were significantly outperformed by a multitask Action Chunking Transformer (ACT)~\cite{zhao2023learning} policy trained from scratch on SutureBot data alone. 
This failure stems from a huge mismatch in both task complexity and physical environment. Unlike most actions in general robot datasets that require minimal instruction, surgical procedures demand years of intensive practice and specialized expertise to execute safely.
Furthermore, surgical instruments interact with deformable, viscoelastic tissues rather than rigid objects; visual feedback relies on complex endoscopic optics rather than standard fixed cameras; and procedures contain multi-step sequences with irreversible actions and narrow tolerance for error. Consequently, most capable general-purpose policies fail at entry-level surgical tasks, not because their architectures are inadequate, but because the gap between general manipulations and surgical procedures is too vast to bridge with out-of-domain pretraining.

The root cause of this domain gap is a structural data bottleneck. Unlike general robotics, acquiring synchronized kinematic and multimodal data in a clinical environment requires navigating stringent safety protocols, institutional review boards, and patient privacy regulations. Instrumenting proprietary surgical robots to record high-fidelity kinematic streams adds further technical complexity rarely encountered in standard robotic data collection. The resulting scarcity is evident in the scale of existing datasets. For over a decade, JIGSAWS~\cite{gao2014jigsaws}, providing merely 3 hours of demonstrations on the da Vinci system, remained the primary surgical manipulation benchmark. Subsequent works have made meaningful progress. The Expanded CRCD~\cite{oh2024crcd} added kinematic recordings of pseudo-cholecystectomy, SutureBot~\cite{haworth2025suturebot} provided around 6 hours of end-to-end suturing data, and ImitateCholec~\cite{Hansen2026ImitateCholec} reached approximately 20 hours of segmented demonstrations of cholecystectomy with paired endoscopic video and dVRK kinematics, making it the largest publicly available single-robot surgical dataset. However, even these important contributions remain orders of magnitude smaller than the massive corpora driving general-purpose manipulation. They are also single-embodiment, single-institution, and narrow in task scope. Prior to the present work, no cross-embodiment surgical dataset has been assembled, leaving the field without the infrastructure that can determine if domain-focused pretraining can overcome the transfer deficit.

Open-H-Embodiment is a direct response to this bottleneck. Inspired by the collaborative success of Open-X-Embodiment~\cite{openxembodiment2024}, but rigorously adapted to the unique sensing, safety, and governance constraints of the medical domain, this community-driven initiative aggregates data from dozens of global institutions. By unifying a large corpus of real and synthetic data across a diverse range of robotic surgery and healthcare tasks, the dataset provides synchronized, multimodal observations under a standardized schema.

Pairing high-capacity VLA models with the Open-H dataset finally allows us to test whether the scaling laws of general robotics apply to the healthcare domain. To investigate this, we introduce GR00T-H, a healthcare-focused foundational VLA developed by post-training the GR00T-N1.6 model on the Open-H dataset. We rigorously evaluate GR00T-H against both generalist and specialist baselines on complex, long-horizon tasks across multiple robotic platforms. Our results show that GR00T-H has significant advantages in overall task success, data efficiency during fine-tuning, and cross-embodiment generalization. Beyond policy learning, we demonstrate that the Open-H enables fine-tuning of world foundation models for surgical robotics simulation. We produce Cosmos-H-Surgical-Simulator (C-H-S-S), the first multi-embodiment, kinematic action-conditioned world model for surgical simulation, built by fine-tuning Cosmos-Predict~2.5~\cite{ali2025world} on the Open-H surgical data mixture spanning nine robotic platforms. C-H-S-S provides a publicly available, commercially usable frontier for in silico policy evaluation and synthetic data generation in the surgical domain. 

In summary, the primary contributions of this work include:

\begin{itemize}
  \item We introduce Open-H, the first large-scale, cross-embodiment, and multimodal dataset for surgical and healthcare robotics.
  \item We present GR00T-H, an open foundational surgical VLA demonstrating superior task success, fine-tuning data efficiency, and cross-embodiment generalization.
  \item We develop C-H-S-S, the first multi-embodiment, kinematic action-conditioned world model for surgical simulation, enabling in silico policy evaluation and synthetic data generation.
\end{itemize}

\section*{Results}
\subsection*{The Open-H-Embodiment Dataset}
\begin{figure*}
  \centering
  \includegraphics[width=0.98\textwidth]{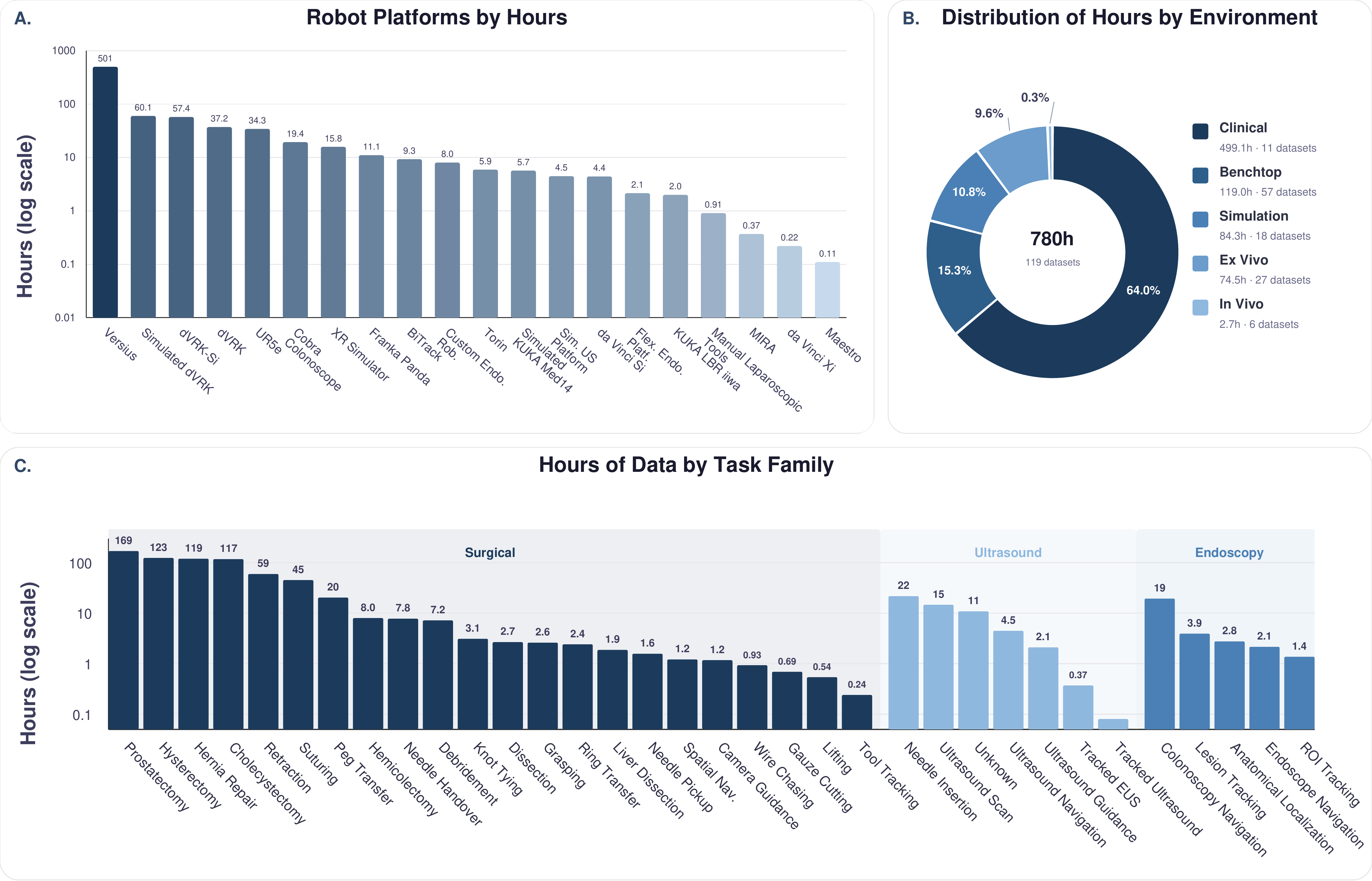}
  \caption{\textbf{Composition of the Open-H-Embodiment dataset.}
    (\textbf{A}) Dataset hours by robot platform.
    (\textbf{B}) Distribution of dataset hours by environment type.
    (\textbf{C}) Distribution of dataset hours across task families.
    Together, these panels summarize the current distribution of contributed data across embodiments, collection environments, and task families in Open-H-Embodiment.}
  \label{fig:dataset_composition}
\end{figure*}

Open-H-Embodiment is the first cross-embodiment, multi-institution dataset for healthcare robotics. The corpus comprises 119 datasets totaling 780 hours of paired video and kinematic data, contributed by more than 50 institutions worldwide. It spans 20 distinct robot platforms, 33 task families, and 5 environment types ranging from digital simulation to live clinical procedures (Figure~\ref{fig:dataset_composition}). Table~\ref{tab:dataset_comparison} contextualizes this scale against prior surgical datasets, all of which are single-embodiment and single-institution.

\begin{table}[b]
\centering
\caption{\textbf{Open-H-Embodiment compared with prior surgical robotics datasets containing kinematic data.} Open-H is the first dataset to span multiple embodiments while exceeding prior work by more than an order of magnitude in duration.}
\label{tab:dataset_comparison}
\footnotesize
\begin{tabular}{lccccc}
\hline
 & \textbf{Hours} & \textbf{Episodes} & \textbf{Platforms} \\
\hline
JIGSAWS \cite{gao2014jigsaws}        & $\sim$3    & 103     & 1 \\
Exp.\ CRCD \cite{oh2024crcd}          & $\sim$7    & 80      & 1 \\
SutureBot \cite{haworth2025suturebot} & $\sim$6    & 1{,}890 & 1 \\
ImitateCholec \cite{Hansen2026ImitateCholec} & $\sim$20 & $\sim$18{,}000 & 1 \\
\textbf{Open-H (ours)}               & \textbf{780} & \textbf{125{,}815} & \textbf{20} \\
\hline
\end{tabular}
\end{table}

\paragraph{Embodiment Diversity:}
The 20 robot platforms span five structural families: surgical robotic systems (618 hours across 76 datasets), industrial arms modified for healthcare use (45 hours, 12 datasets), flexible endoscope robots (30 hours, 8 datasets), simulated robots (86 hours, 21 datasets), and manual instrumentation (1 hour, 2 datasets). Commercial platforms include the Versius (CMR Surgical, Cambridge, UK), da Vinci Si and Xi (Intuitive Surgical, Sunnyvale, USA), BiTrack (Rob Surgical, Barcelona, Spain), MIRA (Virtual Incision, Lincoln, USA), and Maestro (Moon Surgical, Paris, France), alongside research platforms such as the dVRK, KUKA LBR iiwa, Franka Panda, UR5e, and the USTC Torin (Figure~\ref{fig:dataset_composition}a). No prior surgical dataset spans more than a single platform family, making Open-H the first corpus capable of supporting cross-embodiment transfer experiments in this domain.

\paragraph{Task and Procedure Coverage:}
The dataset spans the full granularity spectrum relevant to surgical autonomy, from complete multi-hour clinical procedures down to isolated manipulation primitives. At the procedure level, four clinical workflows each exceed 100 hours: prostatectomy (169 hours), cholecystectomy (172 hours), hysterectomy (122 hours), and hernia repair (119 hours). At the subtask level, the corpus includes suturing and knot tying (57 hours across 30 datasets), tissue manipulation (19 hours, 18 datasets), and skills benchmarks such as peg transfer and needle threading (25 hours, 14 datasets). Beyond surgery, 22 datasets (54 hours) cover robotic ultrasound scanning, tracked ultrasound, and ultrasound-guided intervention. 8 datasets (30 hours) address flexible endoscopy and colonoscopy navigation (Figure~\ref{fig:dataset_composition}c). This coverage across complete procedures and isolated primitives provides training signal for high-level procedural planning and low-level motor control alike.

\paragraph{Sensing Modality Richness:}
Beyond the universal pairing of video and kinematics present in every dataset, individual contributions extend the observation space along multiple sensing axes. 72 datasets (61\%) provide two or more simultaneous camera views, and 52 datasets (44\%) include stereo endoscopic video, providing native depth cues absent from monocular feeds. Wrist-mounted cameras appear in 32 datasets (27\%), offering close-range views of instrument-tissue interaction that complement the endoscopic field of view. Depth sensing is available in 28 datasets (24\%), spanning RGB-D cameras (Intel RealSense, 11 datasets), computed stereo depth, and simulation-derived ground-truth depth maps. The 22 ultrasound datasets pair time-synchronized B-mode imaging with RGB cameras and, in several contributions from CUHK, ImFusion, and TUM, simultaneous wrist-camera and third-person views, creating observation spaces with three or more synchronized visual streams. One ultrasound contribution (TUM CAMP SonATA) additionally provides synchronized force kinematics. The simulation subset (84 hours across 18 datasets) provides dense output modalities infeasible to capture on real tissue, including surface normals, optical flow, and per-pixel tool segmentation masks. Five datasets from UBC include synchronized eye-gaze tracking at 120~Hz, capturing operator visual attention during endoscopic procedures.

\paragraph{Environment and Realism Spectrum:}
The corpus spans five realism tiers: digital simulation (84 hours, 18 datasets), benchtop and phantom (119 hours, 57 datasets), ex vivo (75 hours, 27 datasets), in vivo (3 hours, 6 datasets), and clinical human procedures (499 hours, 11 datasets). Clinical data, predominantly contributed by CMR Surgical (489 hours of Versius procedures), accounts for 64\% of the total corpus by duration (Figure~\ref{fig:dataset_composition}b). This distribution, spanning from fully controlled simulation through cadaveric tissue to live surgery, supports progressive model validation across increasing levels of clinical realism.

\subsection*{GR00T-H: A Foundational Vision-Language-Action Model for Surgery}

To demonstrate the utility of the Open-H dataset in enhancing robot learning performance for surgical tasks, we developed GR00T-H, the first open foundational surgical Vision-Language-Action (VLA) model. GR00T-H is based on NVIDIA's pretrained GR00T-N1.6 policy \cite{nvidia2025groot}, which was then post-trained on the Open-H dataset. We evaluated GR00T-H by comparing its performance against the base GR00T-N1.6, the ACT policy, an architecture previously established as a leading baseline for surgical robotics \cite{haworth2025suturebot}, and LingBot-VA, a recent world action model that we post-train using a 50-hour dVRK-focused subset of Open-H (Supplementary Text). All policies were subsequently fine-tuned on small, embodiment- and task-specific datasets to assess their performance. For the end-to-end suturing experiment, we report cumulative task survival across sequential subtasks. For all other experiments, we report individual sub-task success rates and an averaged sub-task success rate with 95\% Clopper-Pearson confidence intervals and p-values from Fisher's exact test with Holm-Bonferroni correction for multiple comparisons.

\paragraph{End-To-End Suturing:}
To evaluate long-horizon policy performance, we utilize the SutureBot benchmark \cite{haworth2025suturebot}, which employs a dVRK-based suturing dataset with a phantom silicone pad and suture needle as shown in Figure \ref{fig:eval_setups}. To ensure a fair comparison, we implemented a per-setup evaluation protocol: for each specific environment configuration, robot configuration, pad position, and needle placement, GR00T-N1.6, GR00T-H, and ACT each performed a single suturing attempt. This setup was held constant across the three models and only changed once all models had completed the trial. LingBot-VA was evaluated in a separate evaluation session, with a best effort made to mirror the diversity of setup seen by the other three models. The suturing procedure is decomposed into five sub-tasks: needle-pickup, handover, throw, extraction, and knot-tying. Each model was initialized at the first sub-task, and performance was measured by the total progress achieved through the task sequence prior to failure.

\begin{figure}[b!]
    \centering
    \begin{minipage}{0.9\textwidth}
        \centering
        \vspace{6px}
        \includegraphics[width=\textwidth]{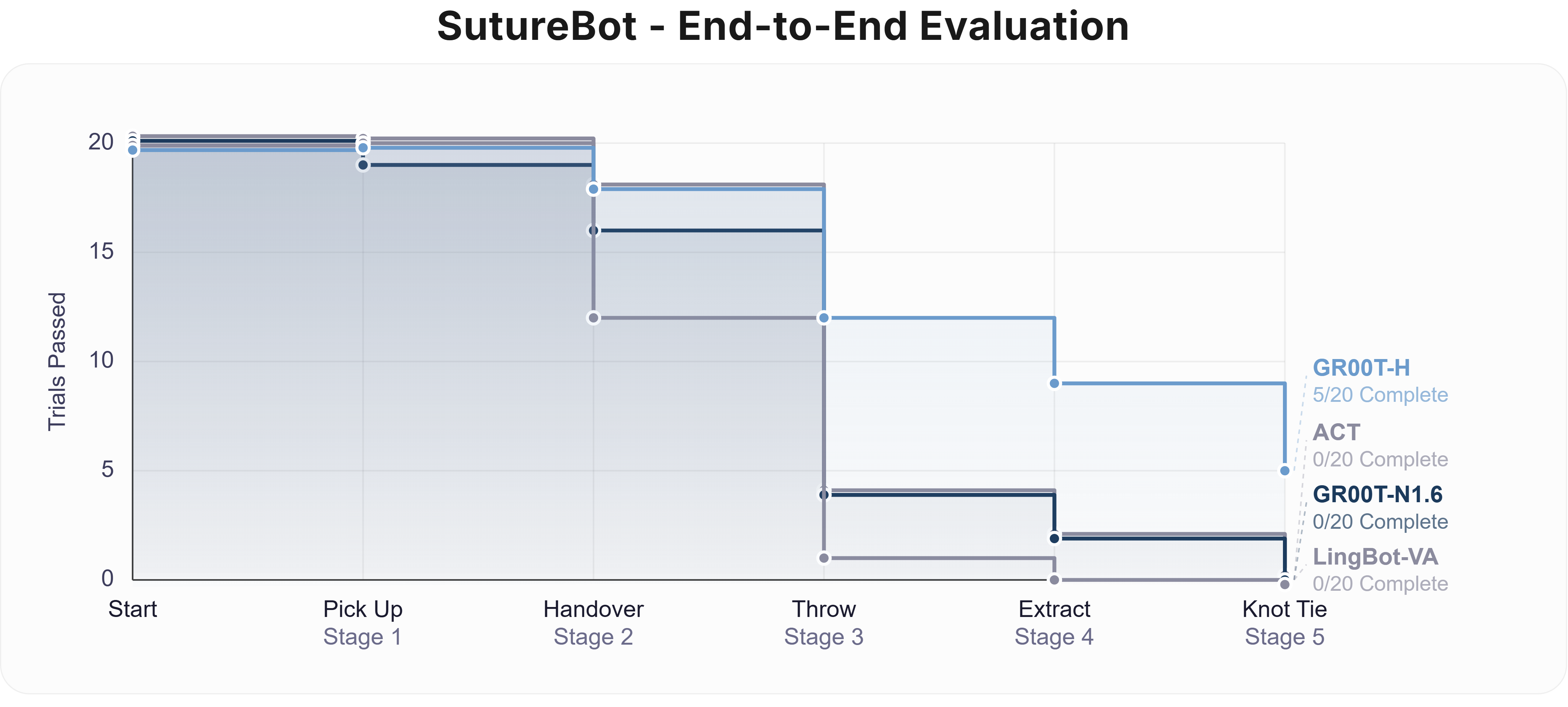}
    \end{minipage}\\ [20pt]
    
    \begin{minipage}{0.9\textwidth}
        \centering
        \includegraphics[width=\textwidth]{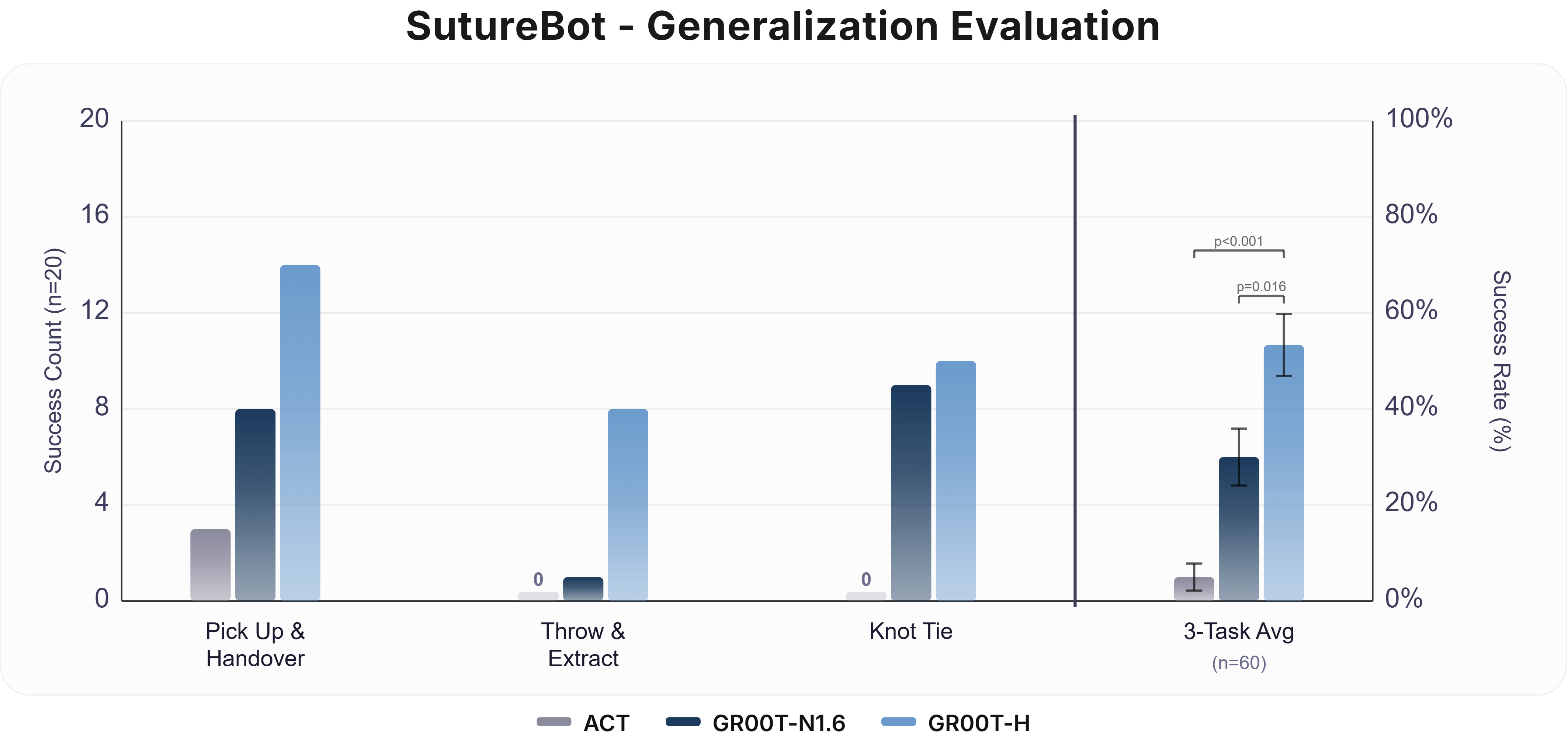}
    \end{minipage}
    \caption{\textbf{SutureBot end-to-end and out-of-distribution evaluation.}
    Top: Task survival for GR00T-H on the SutureBot end-to-end suturing task compared to GR00T-N1.6, ACT, and LingBot-VA. GR00T-H improves end-to-end performance with a long-horizon success rate of 25\%, showing better handling of compounding errors. Bottom: Out-of-distribution evaluation on SutureBot with an unseen wound configuration under varied lighting ($n=20$ per subtask). GR00T-H achieves a 3-task average of 54\%, outperforming GR00T-N1.6 (30\%) and ACT (5\%). Clopper-Pearson 95\% confidence intervals are represented as error bars.}
    \label{fig:end2end_and_generalization}
\end{figure}

As shown in Figure~\ref{fig:end2end_and_generalization}, all four models perform comparably through the early manipulation stages, with GR00T-H retaining all 20 trials through pickup and handover, while by the handover stage GR00T-N1.6 drops to 16, and LingBot-VA falls to 12. The performance gap widens substantially at the throw stage, where GR00T-H retains 12/20 trials compared to 4/20 for both ACT and GR00T-N1.6, and LingBot-VA's 1/20. Only GR00T-H achieves full end-to-end task completion, finishing 5/20 trials (25\%), while ACT, GR00T-N1.6, and LingBot-VA complete 0/20 trials. This experiment emphasizes the importance of policies being robust to compounding errors over the task horizon. A successful end-to-end suturing rollout is shown in \hyperref[fig:gr00t_h_e2e]{Movie~S1}.

\paragraph{Generalization:} 
To assess whether healthcare-specific post-training improves robustness to distribution shift, we evaluated GR00T-H, ACT, and GR00T-N1.6 on a wound configuration unseen during training and under different lighting conditions ($n=10$ trials per subtask, $n=30$ total) as shown in Figure \ref{fig:eval_setups}. Due to relatively low end-to-end performance across policies for the in-distribution setup, we evaluate the policy generalization by looking at average sub-task performance instead of end-to-end performance for a stronger evaluation signal. ACT only achieved 15\% on the pick up and handover task and failed all the other tasks.
Averaging across individual sub-tasks, GR00T-H achieves a success rate of 54\%, compared to 30\% for GR00T-N1.6. Notably, GR00T-H leads on Pick Up and Handover (70\% vs.\ 40\%), and also Throw and Extract (42.5\% vs.\ 5\%). However, GR00T-N1.6 achieved similar performance to GR00T-H on the knot-tying task (50\% vs.\ 45\%). These results confirm that large pretrained VLAs improve robustness and visual generalization, while indicating that GR00T-H post-training may improve robustness in the surgical domain.

\begin{figure}
    \centering
    \begin{minipage}{0.75\textwidth}
        \centering
        \includegraphics[width=\textwidth]{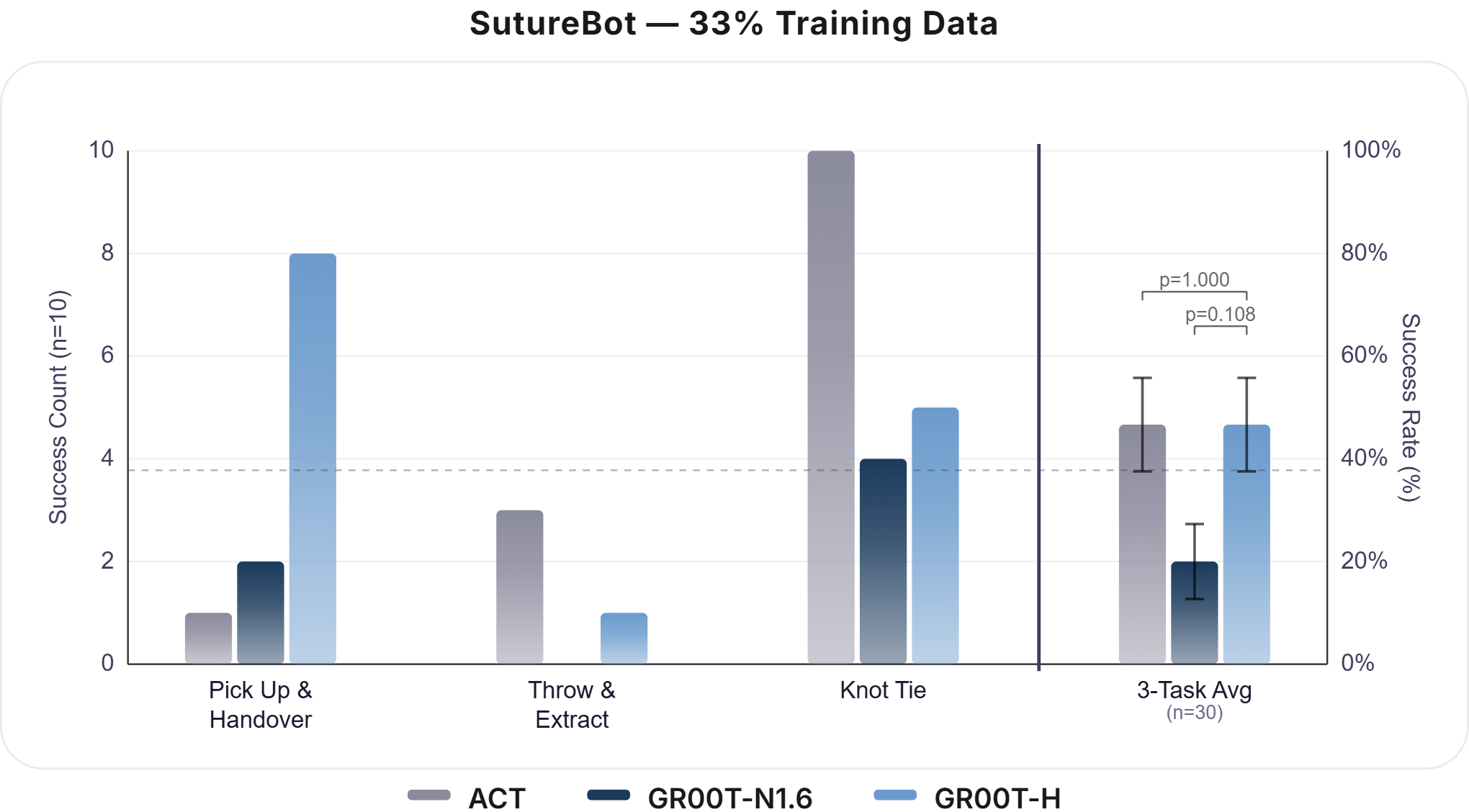}
    \end{minipage}\\ [25pt]
    
    \begin{minipage}{0.75\textwidth}
        \centering
        \includegraphics[width=\textwidth]{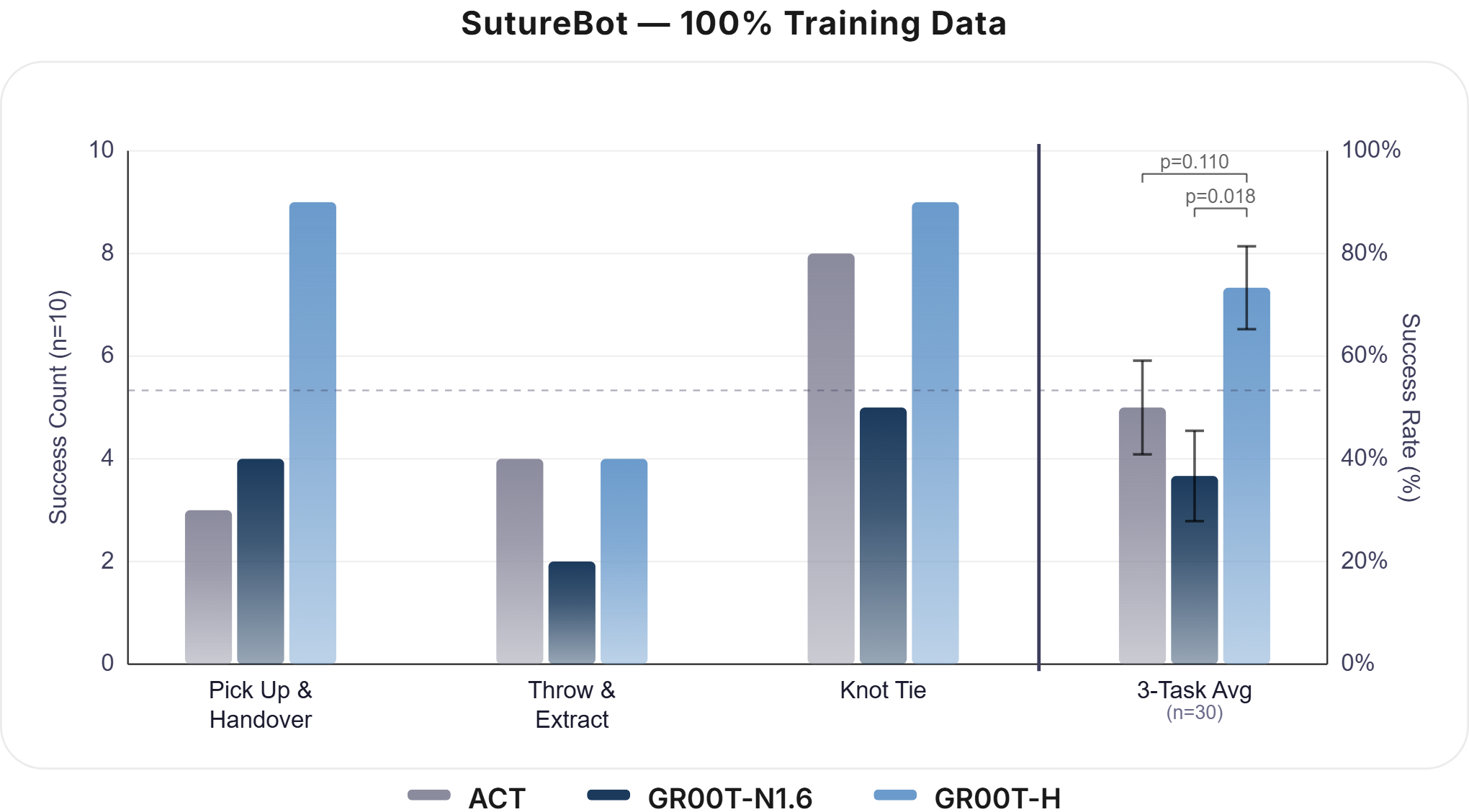}
    \end{minipage}
    \caption{\textbf{SutureBot data efficiency evaluation.}
    Task success rate at 33\% and 100\% fine-tuning data on SutureBot ($n=10$ per subtask). At 33\% data, GR00T-H matches ACT while GR00T-N1.6 underperforms both. At 100\% data, GR00T-H outperforms all baselines, indicating that Open-H post-training enables both data-efficient learning and stronger scaling. Clopper-Pearson 95\% confidence intervals are represented as error bars.}
    \label{fig:data_ablation}
\end{figure}

\paragraph{Data Efficiency:}
To test whether Open-H post-training reduces the amount of task-specific fine-tuning data required, we trained GR00T-H, ACT, and GR00T-N1.6 with either 33\% or 100\% of the full fine-tuning dataset ($\sim$2 hours vs.\ $\sim$6 hours of demonstrations), using proportional sampling to preserve task distribution and the ratio of nominal to recovery demonstrations. Similar to generalization experiments, we use sub-task success rates instead of end-to-end success rates for a stronger evaluation signal. Results are shown in Figure \ref{fig:data_ablation}.

At 33\% data, GR00T-H already matches ACT on the 3-task average (both $\approx$47\%), while GR00T-N1.6 lags at $\approx$20\%. With full data, GR00T-H improves substantially to $\approx$73\%, compared to $\approx$50\% for ACT and $\approx$37\% for GR00T-N1.6. This suggests that Open-H post-training provides a stronger initialization for surgical fine-tuning: GR00T-H achieves competitive performance with limited 
data and may scale more effectively than GR00T-N1.6 when more data is available.

\begin{figure}
    \centering
    \includegraphics[width=1\linewidth]{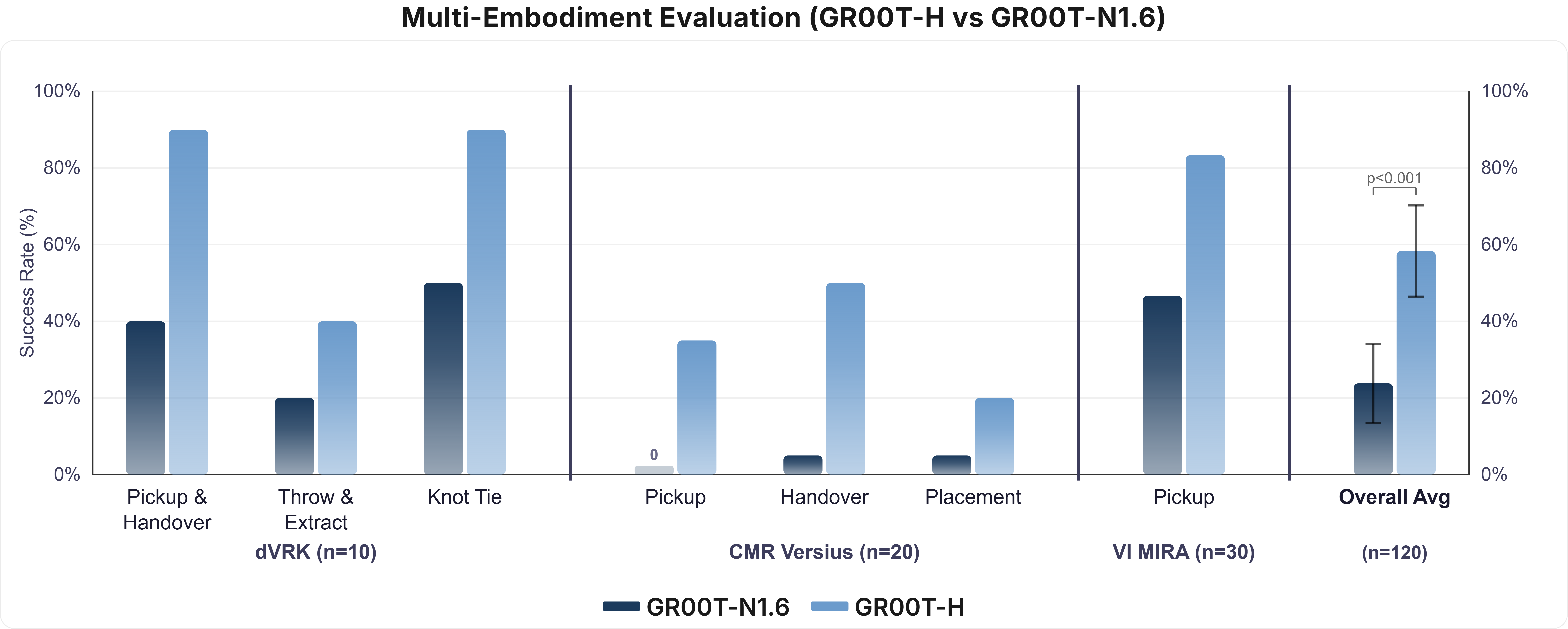}
    \caption{\textbf{Multi-Embodiment Performance Comparison.}
    Evaluation of the GR00T-H foundational VLA (post-trained on Open-H) versus the GR00T-N1.6 base policy across three surgical platforms: the da Vinci Research Kit Si (dVRK-Si), CMR Versius, and Virtual Incision MIRA. GR00T-H demonstrates significant performance gains across all robot embodiments and sub-tasks, with the overall average success rate showing a statistically significant improvement ($p < 0.001$). Error bars represent the Clopper-Pearson 95\% confidence intervals across trials.}
    \label{fig:multiembodiment}
\end{figure}

\paragraph{Multi-Embodiment:}
A primary advantage of training VLAs on diverse multi-embodiment data is the resulting performance gain across distinct platforms within the data distribution. To validate GR00T-H’s cross-embodiment capabilities, we compared it against the GR00T-N1.6 base model across three systems: the CMR Versius, Virtual Incision MIRA, and the da Vinci Research Kit Si (dVRK-Si). The Versius system was evaluated on Peg Transfer, a standard surgical training task, using 5.2 hours of training data, with the procedure decomposed into block pickup from the peg, block handover, and placement on a peg. For the MIRA robot, we evaluated the needle pickup sub-task using only 22 minutes of data. Finally, the dVRK-Si was used for the SutureBot sub-tasks described previously, trained on 6 hours of data. Images of all system setups can be found in Figure \ref{fig:eval_setups}. As shown in Figure \ref{fig:multiembodiment}, GR00T-H demonstrates a significant performance boost over the base model across all platforms ($p < 0.001$ for the overall average), confirming that Open-H post-training effectively enhances surgical capabilities across diverse robotic embodiments.

\begin{figure}
    \centering
    \includegraphics[width=1.0\linewidth]{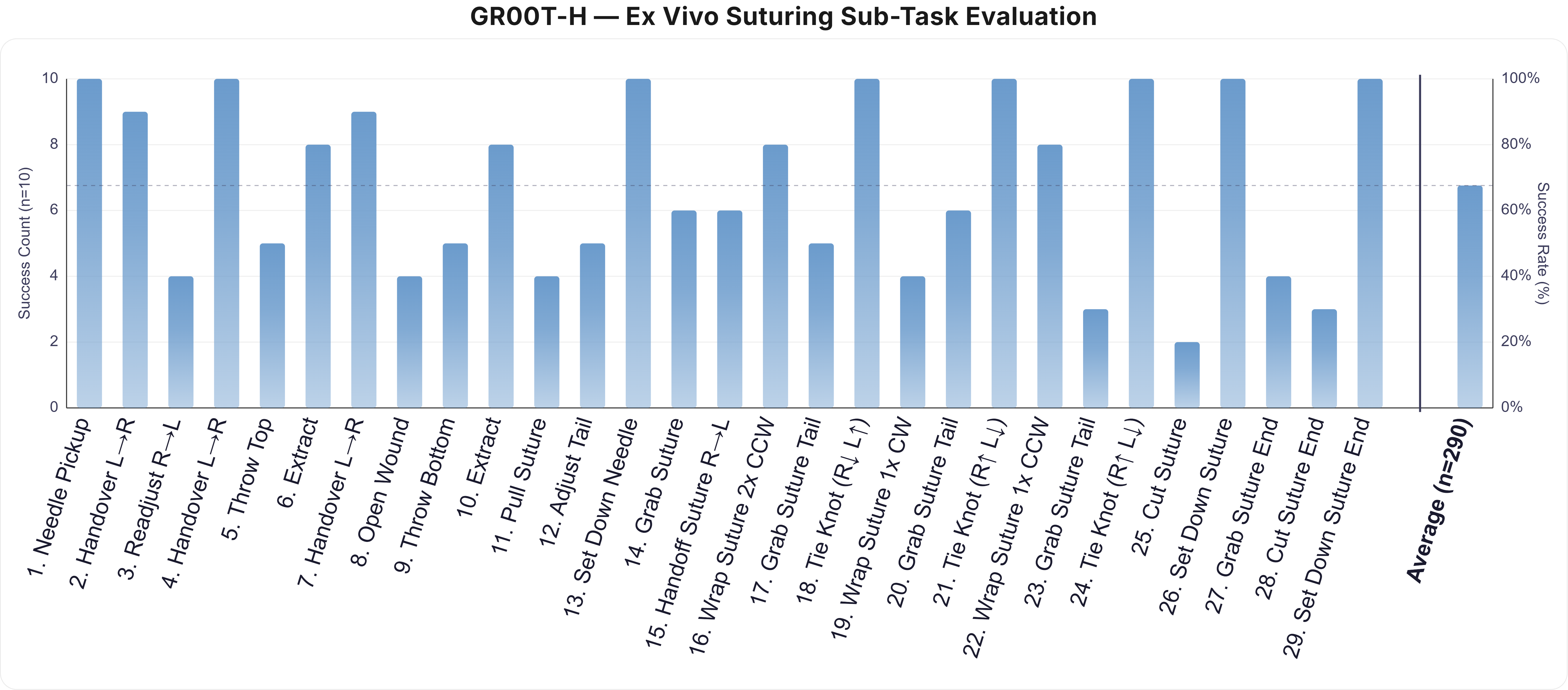}
    \caption{\textbf{Ex vivo suturing evaluation across 29 subtasks.} 
    ($n=10$ per subtask, $n=290$ total). Tasks span needle manipulation, wound opening, suture passing, knot tying, and suture cutting. GR00T-H achieves an average success rate of $\approx$64\%, with near-perfect performance on structured manipulation primitives and lower success on fine-contact and cutting steps. The rightmost bar shows the overall average across all subtasks. Clopper-Pearson 95\% confidence intervals are represented as error bars.}  
    \label{fig:ex_vivo}
\end{figure}

\paragraph{Ex Vivo Experiment:}
To evaluate GR00T-H in a clinically proximate setting, we conducted an ex vivo suturing evaluation on skin-on pork belly, assessing performance across the full 29-subtask sequence required to complete a suture covering needle manipulation, needle throwing, wound opening, suture manipulation, knot tying, and suture cutting. The ex vivo suturing setup can be seen in Figure \ref{fig:eval_setups}. Each subtask was evaluated over 10 trials ($n=290$ total), with final results shown in Figure~\ref{fig:ex_vivo} and video of successful subtask rollouts shown in \hyperref[fig:gr00t_h_wound_closure]{Movie~S2}.

GR00T-H achieves an overall average success rate of 64\% across all 29 subtasks. Performance is strongest on structured manipulation primitives: needle pickup (10/10), all handover stages (9/10, 10/10, 9/10), set-down needle (10/10), and all three knot-tying steps (10/10 each) all reach near-perfect or perfect success. Performance is lower on subtasks requiring fine instrument coordination or tissue 
contact, including readjust (4/10), open wound (4/10), wrapping steps (8/10, 4/10, and 8/10), grab the suture tail (5/10, 6/10, and 3/10), and the two cut suture steps (2/10 and 3/10). The cutting steps represent the most consistent failure mode, potentially due to the precise and quick nature of the cutting motion.  

These results demonstrate that GR00T-H can execute the sub-tasks of a clinically relevant manipulation sequence on real biological tissue, with reliable performance on the majority of sub-tasks. The pattern of failures, concentrated in fine-contact and cutting steps rather than distributed uniformly, suggests that targeted data collection and fine-tuning on these specific failure modes represent a tractable path to achieve end-to-end performance.

\begin{figure}
  \centering
  \includegraphics[width=1.0\textwidth]{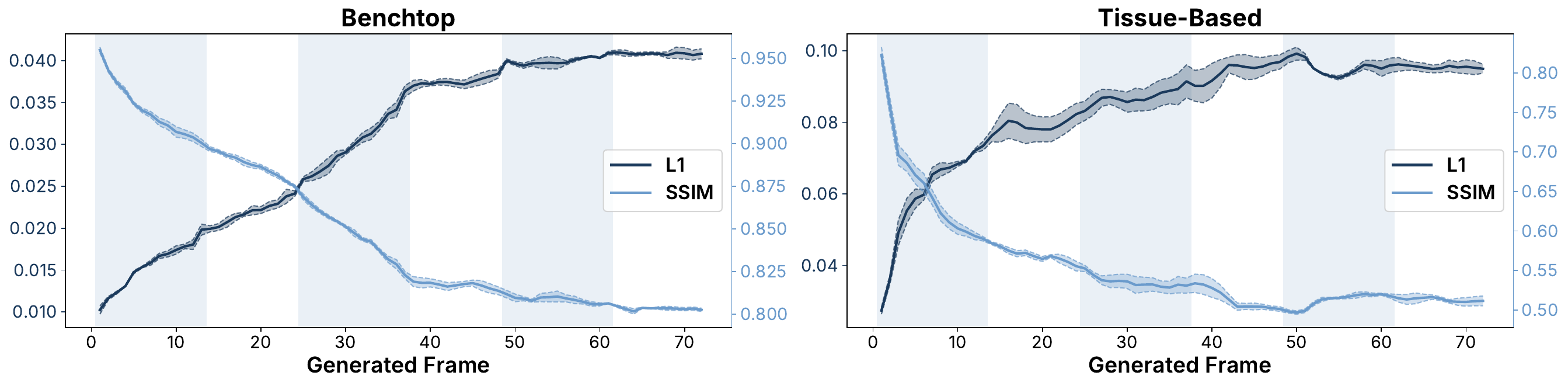}
  \caption{\textbf{Quantitative evaluation of Cosmos-H-Surgical-Simulator.}
    Per-frame L1 and SSIM for benchtop vs. tissue-based datasets. Mean L1 error and SSIM as a function of generated frame index across 72 autoregressively generated frames (6 chunks × 12 frames each). Mean over 3 generation seeds, each averaged across all evaluated episodes within the category; shaded bands indicate 1 standard deviation across seeds. Left: benchtop datasets (phantom and bench procedures). Right: tissue-based datasets (clinical, cadaver, and ex vivo tissue).}
  \label{fig:cosmos_h_quant_results}
\end{figure}

\subsection*{Cosmos-H-Surgical-Simulator: A World Model for Surgical Robotics Simulation}
Cosmos-H-Surgical-Simulator is the first multi-embodiment, kinematic action-conditioned world foundation model for surgical robotics simulation. Built by fine-tuning Cosmos-Predict~2.5~\cite{ali2025world}, a 2B-parameter latent video diffusion transformer, on the Open-H surgical data mixture spanning nine robotic platforms and 32 datasets (Table~\ref{tab:chss-dataset-mixture}), C-H-S-S accepts a single video frame and a sequence of kinematic action vectors, and autoregressively generates future frames that predict the visual outcome of those actions. Through iterative rollout, it can produce complete surgical trajectory videos for any embodiment in its training distribution, enabling in silico policy evaluation and synthetic data generation from a single publicly available, commercially usable checkpoint. \hyperref[fig:c_h_s_s_viz]{Movie~S3} shows representative generated rollouts across multiple embodiments. To our knowledge, no prior surgical world model has supported more than a single robotic platform.

Quantitative evaluation of surgical world models remains an open problem. Standard unconditional video generation metrics such as FID and FVD measure distributional similarity to a reference set but do not assess whether a generated video faithfully tracks the specific visual consequences of a given action sequence. Because C-H-S-S is action-conditioned, replaying the recorded action sequence from a held-out test episode produces a generated rollout that can be compared frame-by-frame against the original ground-truth video. We adopt two complementary pixel-level metrics: L1 (mean absolute error), which measures overall pixel fidelity, and SSIM (structural similarity index), which captures perceptual structural correspondence. Both are computed per frame across 72 autoregressively generated frames (6~chunks of 12~frames each) on held-out test episodes from 25 Open-H datasets, using 3 generation seeds per episode to quantify variability. Results are stratified by dataset category; benchtop (datasets covering phantom and bench procedures) and tissue-based (datasets covering clinical, cadaver, and ex vivo tissue), to characterize how scene complexity affects generation quality.

Figure~\ref{fig:cosmos_h_quant_results} reports the per-frame L1 and SSIM trajectories for both categories. Benchtop scenes, which feature controlled lighting, static backgrounds, and clearly visible instruments, maintain low L1 and high SSIM throughout the 72-frame horizon, with only gradual degradation. Tissue-based scenes exhibit higher L1 and lower SSIM overall, consistent with the greater visual complexity of in-body environments: variable lighting from the endoscope, frequent instrument occlusion, specular reflections on wet tissue, and deformable anatomy that is inherently harder to predict. The error bands across seeds are narrow for benchtop scenes, indicating stable generation; tissue-based scenes show wider seed-to-seed variability, particularly in L1, reflecting the greater stochasticity involved in predicting visually complex in-body environments. The expected sawtooth pattern at chunk boundaries, where each new 12-frame chunk re-conditions on the last generated frame, is visible but modest, suggesting that the model maintains coherent scene state across autoregressive transitions.

\section*{Discussion}
Our results show that training on a large, diverse surgical dataset improves both task performance and robustness of foundation VLA policies. GR00T-H was the only model to achieve full end-to-end task completion on SutureBot, and it consistently outperformed ACT and the base GR00T-N1.6 model under out-of-distribution conditions and with limited fine-tuning data. These findings suggest that the scale and diversity of Open-H provide a surgical 
prior that transfers across tasks, embodiments, and environments. The data bottleneck constraining surgical robot learning is not fully resolved, but 
it can be meaningfully reduced through large-scale data collection paired with healthcare-specific foundation model post-training.

Our evaluation also revealed an interesting finding about policy robustness to hardware variation. Over the course of the evaluation period, the experimental 
setup underwent changes typical of long-running robot deployments: surgical instruments accumulated mechanical wear that shifted their cable-driven kinematics, and wrist cameras required replacement with units of slightly different optical properties. ACT was sensitive to these changes and performed below its originally reported results~\cite{haworth2025suturebot}. GR00T-H, on the other hand, still achieved 25\% end-to-end task completion under the same conditions. 
This suggests that post-training on the diverse Open-H dataset makes the policy more resilient to the kind of gradual hardware drift that is common in real 
surgical environments, where instrument wear and camera replacements are routine.

Post-training on Open-H also appears to improve data efficiency. GR00T-H matches ACT with only 33\% of the full fine-tuning dataset and outperforms the GR00T-N1.6 baseline at 100\%. This suggests that post-training on Open-H provides a stronger starting point for learning new surgical tasks, which is useful for institutions that want to adapt the model to new platforms or procedures without collecting large amounts of local data. The multi-embodiment results further support this interpretation: GR00T-H shows consistent gains across different robotic platforms, suggesting the learned representations are not specific to a single robot's kinematics.

Open-H also enables a new class of world foundation model for surgical simulation. Cosmos-H-Surgical-Simulator is, to our knowledge, the first kinematic action-conditioned world model that supports multi-embodiment surgical video generation from a single checkpoint, spanning nine robotic platforms. Prior surgical world models~\cite{cosmos_surg_dvrk} have been limited to a single platform; C-H-S-S extends this to any embodiment in its training distribution without platform-specific retraining. The per-frame L1 and SSIM evaluation confirms that the model maintains visual fidelity over 72-frame autoregressive rollouts, with the performance gap between benchtop and tissue-based scenes tracking the inherent difficulty of the underlying environments. Prior work on the dVRK platform~\cite{cosmos_surg_dvrk} has further demonstrated that action-conditioned world models of this kind can serve as simulation environments for closed-loop policy evaluation. Open-H can therefore support not just policy training but also world models, visual encoders for instrument and tissue understanding, and language-conditioned planning systems, making it useful as shared infrastructure for the broader surgical robotics community.

Several limitations of the current work must be addressed before these methods could be applied in animal or human surgical settings. In ex vivo evaluation, GR00T-H performs well on structured steps like needle pickup, handover, and knot tying, reaching near-perfect success on many of these. However, performance drops on subtasks requiring fine instrument coordination under tissue contact, with cutting steps reaching only 20--30\% success. The overall average of 64\% across the full 29-step ex vivo suturing sequence, and 25\% end-to-end completion on the SutureBot benchmark, are encouraging but well below what would be needed for reliable autonomous execution. All evaluations are conducted on tissue phantoms and benchtop ex vivo tissue in controlled lab settings. How the policy performs on live tissue, which varies in stiffness, bleeds, and moves, is not yet known. The policy also has no way to detect or respond to unexpected events such as tissue tearing, instrument failures, or patient movement, all of which are important safety considerations for any real deployment. Errors also tend to compound across long task sequences, which limits end-to-end completion rates even when individual tasks are performed reliably. 
More data from a wider range of institutions, procedures, and tissue types will be needed for the model to generalize to the variety of cases seen in clinical practice. While initial procedural annotations covering approximately 205 hours of Versius-500 clinical data are released alongside the dataset, richer labels including instrument state, tissue contact events, and failure modes will be needed to address the fine-contact steps where the model currently struggles the most. Pre-clinical evaluation in animal models, with appropriate regulatory oversight, is a necessary next step before any component of this pipeline could be considered for use in humans.
LingBot-VA, which leverages a world foundation model backbone as opposed to a VLM backbone, represents a promising but distinct architectural class for robot learning. Its comparatively rapid performance degradation across subsequent end-to-end suturing subtasks does not necessarily reflect the potential of the architecture, as several factors remain unexplored. The model was post-trained on a 50-hour dVRK-focused subset of Open-H, roughly an order of magnitude less data than GR00T-H's 601-hour surgical mixture. World action models also introduce extensive training and inference parameters, including action and video denoising schedules, KV cache update frequency, video frame rate, and context history length, all of which require further exploration to properly adapt these models to surgical tasks. Determining the appropriate data recipe and inference configuration for world action models on dexterous surgical data is an important direction for future work.
Regarding C-H-S-S, the L1 and SSIM evaluation relies on open-loop replay of recorded action sequences. In closed-loop deployment, where actions come from a policy or a surgeon's console reacting to the model's own generated frames, no such ground truth exists, and pixel-level metrics are no longer applicable. More broadly, L1 and SSIM are scene-level measures that do not capture surgery-specific aspects of generation quality, such as whether instruments are rendered in the correct position or whether tool-tissue interactions are physically plausible. Developing domain-specific metrics, such as spatial consistency of generated instruments against ground-truth tool trajectories or fidelity of tool-tip localization across frames, is an important open problem. Early experiments with segmentation-based metrics such as generated-vs-actual tool consistency and tool centroid distance showed promise on individual platforms but proved insufficiently robust across the diversity of embodiments and visual conditions. Additionally, the Open-H dataset predominantly comprises successful task demonstrations. Prior work has shown that failure episodes are a critical factor for surgical world model training~\cite{cosmos_surg_dvrk}. Future data collection efforts should explicitly include such failure trajectories to improve the realism and coverage of surgical world models. Extending the automated closed-loop evaluation pipeline demonstrated for a single platform~\cite{cosmos_surg_dvrk} to the multi-embodiment setting of C-H-S-S is an important next step.

\section*{Materials and Methods}
\subsection*{The Open-H Dataset}

While large-scale, multi-embodiment datasets have driven rapid progress in general-purpose robotics, assembling a comparable corpus for healthcare robotics introduces challenges that are largely absent from that setting.    

In general-purpose robotics, foundational policy research is predominantly conducted on purpose-built platforms such as the Franka Panda~\cite{kim2024openvla}, Unitree G1~\cite{nvidia2025groot}, and ALOHA~\cite{black2024pi0}. These platforms provide integrated hardware and software that support teleoperation and data collection with minimal additional engineering. In healthcare robotics, by contrast, research is largely conducted on devices engineered for clinical use rather than for robotics research. Most contributing institutions have therefore developed their own suites of proprietary hardware, software, and interfaces to repurpose these medical devices for modern robotics workflows.

Creating a multi-institution, multi-embodiment healthcare-robotics dataset that is \emph{usable} by downstream researchers is constrained by two principal challenges:
\begin{enumerate}
  \item \textbf{Inconsistent data formatting.} Training downstream models on dozens of independently formatted datasets requires detailed knowledge of each collection's schema to convert the complete corpus into a unified representation.
  \item \textbf{Opaque platform nuances.} The diversity of hardware and task families in healthcare robotics introduces considerations that are not self-evident from the data alone, such as surgical clutching mechanisms that invalidate portions of recorded kinematics, multi-arm systems in which only a subset of visible instruments are actively controlled, and variation in operator skill level (e.g., expert surgeon versus machine-learning researcher).
\end{enumerate}

To address these concerns, Open-H adopts two complementary strategies: all constituent datasets are converted into a unified LeRobot v2.1 format, and each dataset is accompanied by a consistently structured README that discloses platform-specific nuances to downstream users.

\paragraph*{Data Formatting:}
We adopt the LeRobot v2.1 format as the standard representation for all Open-H data on the basis of its storage efficiency and broad community adoption~\cite{cadene2026lerobot}. The format stores low-dimensional kinematics in columnar Parquet files and visual observations as hardware-accelerated, lossy MP4 video, reducing the overall storage footprint by up to 98\% compared to formats such as RLDS or HDF5~\cite{chen2025robodm}. Healthcare-robotics tasks, however, frequently require contextual metadata that falls outside the standard LeRobot schema for video, kinematics, and task prompts, including modality-specific parameters (e.g., ultrasound acquisition settings), task-level annotations (e.g., surgical phase and active instrument identity), and other domain-relevant fields. To accommodate these requirements without breaking compatibility with the core LeRobot library, we developed the Open-H data-collection repository~\cite{openh_embodiment_repo}, which provides standardized guidelines and examples for extending the LeRobot schema with healthcare-specific fields.

\paragraph*{Documentation:}
Even with a unified storage format, the practical utility of an aggregate dataset at this scale depends on whether downstream users can understand the collection-specific details of each constituent contribution. To this end, every Open-H dataset includes a structured README that follows a common template. The template captures the robot type(s) used, the data-collection method (e.g., teleoperation or autonomous execution), the operator skill level (e.g., expert surgeon or machine-learning researcher), the data synchronization strategy employed, the dimensions of data diversity (e.g., camera positioning, environment, and lighting variation), the kinematic representation (e.g., absolute Cartesian or relative joint space), and additional domain-relevant fields. The complete template with all recorded fields is available in \cite{openh_embodiment_repo}.

\paragraph*{Clinical Procedure Annotation:}
The clinical procedures in the Versius-500 contribution present a distinct challenge for downstream learning. Unlike tabletop manipulation tasks, where the correct next action is largely determined by a single observation frame, clinical surgical procedures are temporally extended and non-Markovian: from a given endoscopic image, the next action is not immediately clear, as it depends on temporal context, clinical and patient history, as well as surgeon preference. To support future research that leverages this structure, we release initial annotations alongside the dataset. For Versius-500-inguinal-hernia, 13 episodes covering approximately 32 minutes of procedure video were manually annotated at the pixel and frame level (5~FPS) with instrument and anatomy segmentation masks and action classifications. For Versius-500-cholecystectomy, we manually annotated 500 minutes of phase-level and gesture-level labels, and for Versius-500-hysterectomy, 1000 minutes phase-level labels. We fine-tuned LemonFM~\cite{che2026lemon}, a surgical video foundation model pretrained on 938 hours of endoscopic video, on this manually annotated data, and generated labels for the full 80 hours of Versius-500-cholecystectomy and 124 hours of Versius-500-hysterectomy. Evaluated against a held-out test subset of Versius-500 data, the model achieves phase accuracy of approximately 0.65 and gesture accuracy of 0.56 on cholecystectomy, and phase accuracy of 0.75 on hysterectomy.

\subsection*{GR00T-H}
The Open-H dataset supports a range of downstream applications, from monocular depth estimation models to world models, though its most immediate use is as a pre-training corpus for healthcare-focused manipulation policies. To test whether domain-specific data at this scale can close the transfer gap documented above, we train GR00T-H, a surgical foundation policy built on the GR00T-N1.6-3B vision-language-action model (VLA).

We build GR00T-H on GR00T-N1.6-3B~\cite{nvidia2025groot} due to the strong foundation it establishes for training an imitation learning policy on the multimodal and multi-embodiment Open-H corpus. The model's extensive pre-training on diverse real and synthetic manipulation data provides a meaningful initialization for the dexterous, long-horizon tasks common in surgical robotics. Its Cosmos-2B VLM backbone encodes images at flexible resolution, accommodating the heterogeneous resolutions and aspect ratios across healthcare-specific robotic embodiments. Furthermore, its embodiment-specific action heads allow each robot configuration to learn its own output projection, an important feature when training across diverse robot embodiments with varying kinematic output distributions.

GR00T-H is adapted from the upstream foundation model using an Open-H post-training phase. For this post-training, we select a 601-hour surgical subset of the full 780-hour Open-H corpus, isolating the largest and most coherent subgroup: real-world surgical tasks. Thus, only real-world surgical datasets were used in the training of GR00T-H; we leave training and evaluating policies that generalize across ultrasound, endoscopy, and simulation data for future work. For the sampled real-world surgical subset, the Versius-500 contribution dominates by volume. Therefore, we cap its sampling frequency at 20\% of training steps to prevent any single embodiment, environment, or task family from dominating the loss signal. The remaining datasets are sampled proportionally to their size in the corpus. The full training mixture is provided in Table~\ref{tab:gr00t-h-dataset-mixture}.

To establish a common kinematic learning space across embodiments, we normalize all datasets to use relative end-effector (EEF) control with 6D rotation matrices~\cite{zhou2019continuity} for the orientation component. Relative actions remove the requirement for the model to learn each unique embodiment's forward kinematics, a necessity under joint-space control, and improve robustness to variation in workspace position and scale, following the relative action formulation adopted in prior surgical policy work~\cite{kim2024srt}.

To accommodate the heterogeneous kinematics across the Open-H corpus, we leverage GR00T-N1.6's embodiment-specific action heads, where each unique robot configuration learns its own MLP projection from the DiT action expert outputs to the normalized action space. We assign a distinct action head to each robot configuration rather than sharing heads across datasets collected on the same platform. This accommodates a practical constraint of cable-driven surgical robots: cable stretch, hysteresis, and wear characteristics vary across individual machines, and different instrument types introduce further variation in tool-tip kinematic mapping. A shared projection would conflate these instance-specific differences, degrading action prediction accuracy.

To ensure well-normalized kinematic values across robot configurations, we collect per-dataset normalization statistics and merge them into per-configuration statistics at train time via weighted moment matching, where the weights correspond to training-mixture sampling ratios. Statistics are computed per action dimension and per temporal step within the action chunk to preserve resolution for near-future actions~\cite{lbmtri2025}. We apply z-score normalization with $[-5, 5]$ clipping to faithfully model each configuration's kinematic distribution while excluding extreme outliers.

The design decisions described above, including the data mixture, action normalization scheme, embodiment-specific action heads, and kinematic normalization strategy, were iteratively refined using a combination of real-world evaluation and Cosmos-Surg-dVRK~\cite{cosmos_surg_dvrk}, an automated evaluation pipeline. Validation loss on held-out demonstrations is a poor proxy for closed-loop task success in imitation learning, making standard early-stopping criteria unreliable for checkpoint selection. Cosmos-Surg-dVRK addresses this by using an action-conditioned world model (architecturally identical to C-H-S-S, trained on monocular SutureBot data) to autoregressively render full task-length rollouts from the policy under evaluation, after which a finetuned V-JEPA~2~\cite{assran2025vjepa2} labels the generated video with per-subtask success rates. Prior work established a strong Pearson correlation between this pipeline's automated scores and real-world success rates on the dVRK~\cite{cosmos_surg_dvrk}. During GR00T-H development, Cosmos-Surg-dVRK served as a practical screening tool to narrow the space of checkpoint iterations and training hyperparameters before committing physical robot time. 

Using this iteratively validated recipe, the final Open-H post-training run takes 65{,}000 steps of full-weight training with a global batch size of 1{,}024 on the 601-hour surgical subset. The complete post-training configuration, including optimizer, learning rate schedule, and augmentation parameters, is provided in Table~\ref{tab:gr00t_h_mid_training_details}. For downstream use, including all evaluations in this work, we recommend an additional fine-tuning stage in which the VLM backbone is frozen and only the action prediction components are fine-tuned on the target task, following the protocol established in GR00T-N1.6~\cite{nvidia2025groot}. This preserves the healthcare-domain representations acquired during post-training while efficiently specializing the action expert DiT to the target task distribution.

\subsection*{Cosmos-H-Surgical-Simulator}
Beyond policy learning, the Open-H dataset enables fine-tuning of world foundation models for surgical robotics simulation. Kinematic action-conditioned video generation models that predict future visual observations given a current frame and a sequence of kinematic actions are a promising frontier for in silico surgical policy evaluation and synthetic data generation (SDG). By learning the visual dynamics of instrument-tissue interaction, such models can provide a low-cost evaluation environment for surgical policies before physical deployment, as previously demonstrated for the dVRK platform~\cite{cosmos_surg_dvrk} and in the development of GR00T-H. We extend this paradigm to the multi-embodiment setting by fine-tuning Cosmos-Predict~2.5~\cite{ali2025world}, a 2B-parameter latent video diffusion transformer, on the surgical subset of Open-H, producing the Cosmos-H-Surgical-Simulator checkpoint. Cosmos-Predict~2.5 is a diffusion transformer (DiT) that generates video in the latent space of a pre-trained variational autoencoder. An MLP conditions the denoiser on per-timestep kinematic actions, enabling action-conditioned next-frame prediction. Given a single context frame and a sequence of 12 action vectors, the model generates 12 subsequent frames. Through autoregressive rollout, the model produces videos covering complete surgical trajectories. We adopt the publicly released Cosmos-Predict~2.5-2B-Video2World weights as initialization and fine-tune on Open-H surgical data with a unified 44-dimensional action space that accommodates all embodiments via zero-padding.

The C-H-S-S training mixture comprises 32 datasets from 9 robot embodiments and 10+ institutions. CMR Surgical Versius-500 accounts for 50\% of the training compute across four clinical procedures (cholecystectomy, prostatectomy, inguinal hernia, hysterectomy), while the remaining 50\% is distributed proportionally by frame count across dVRK variants (JHU, Stanford, Hamlyn, UCSD, UCB), Turin MITIC, USTC Torin, and Moon Surgical platforms. All action representations are converted to a hybrid-relative format (relative translation + 6D rotation) and z-score normalized per-embodiment. The full dataset specification and mixture ratios are detailed in Table~\ref{tab:chss-dataset-mixture}.

Fine-tuning uses fused AdamW with a learning rate of $1.6 \times 10^{-4}$, linear decay schedule with warm-up, and a global batch size of 1{,}024, running for 42{,}000 steps on 64 A100 80\,GB GPUs. Video resolution is $512 \times 288$ (16:9). The model processes 13 frames per sample (1 context + 12 prediction) with 12 corresponding action timesteps. Each embodiment's native kinematic frequency is downsampled to a consistent $\sim$10\,fps effective rate via embodiment-specific timestep strides.

To quantitatively assess generation fidelity, we replay recorded action sequences from held-out test episodes through C-H-S-S and compare the generated video frame-by-frame against the ground-truth recording. Each episode is rolled out for 6 autoregressive chunks (72 generated frames). We report two pixel-level metrics: L1 (mean absolute error in $[0,1]$ pixel space), which measures overall pixel fidelity, and SSIM (structural similarity index), which captures perceptual structural correspondence. Evaluation covers 25 of the 32 training datasets (7 are excluded because their test-split episodes are shorter than the 72-frame evaluation horizon). For each dataset, 2 episodes are drawn from the 5\% held-out test split using a fixed selection seed, and each episode is generated with 3 independent random seeds, yielding up to 150 episode evaluations in total. To isolate generation variability from cross-dataset differences, we aggregate by first computing the per-frame mean across all episodes within each seed, then reporting the mean and standard deviation across the 3 seed-level means. Results are stratified into benchtop (18 datasets: bench procedures) and tissue-based (7 datasets: clinical, cadaver, and ex vivo tissue) categories.


\clearpage

\bibliography{references}
\bibliographystyle{sciencemag}


\section*{Acknowledgments}
We would like to thank the many institutions, companies, researchers, and students who participated in this effort; without their contributions, this dataset would not have been possible.

\paragraph*{Funding:}
This material is based on work supported by the following funding sources: NIH R56EB033807 (A.K.), ARPA-H award 75N91023C00048 (J.C., Xin.C., A.K.), ARPA-H award AY1AX000023 (J.G., E.K., A.K.), ARPA-H award D24AC00415-00 (N.Y., A.K.), NSF CAREER Award 2144348 (A.K.), and NSF Graduate Research Fellowship 2023354859 (Je.H.). Pe.K. and Hao.Z. were partially supported by NSF AccelNet awards 1927354 (JHU) and 1927275 (WPI). A.M.O.'s contribution was supported by the Stanford Institute for Human-Centered Artificial Intelligence and the Stanford Robotics Center. A.E.A.'s contribution was supported by the Stanford Institute for Human-Centered Artificial Intelligence. T.H., E.L. and K.T. were partially supported by Project 2024-1.2.3-HU-RIZONT-00069, implemented with the support provided by the Ministry of Culture and Innovation of Hungary from the National Research, Development and Innovation Fund, financed under the 2024-1.2.3-HU-RIZONT funding scheme. T.H. is a consolidator researcher, supported by the Distinguished Researcher Program of \'Obuda University. Zh.C., Yam.Z., T.Y., and Yun.T. were funded by the Multi-scale Medical Robotics Center, AIR@InnoHK. Xia.C. was funded by the Multi-scale Medical Robotics Center, AIR@InnoHK and Direct Grants (The Chinese University of Hong Kong) under Grant 4055245. P.V., D.J., B.C., and Ju.H. were supported by the Engineering and Physical Sciences Research Council (EPSRC) under Grants EP/V047914/1 and UKRI180 and by the National Institute for Health and Care Research (NIHR) Leeds Biomedical Research Centre (BRC) (NIHR203331). J.L., M.S., G.W., Zi.H., E.W., and H.R. were supported by the NSFC Young Scientists Fund - Scheme A T252500134, the Ministry of Science and Technology (MOST) of China Key Project 2025YFE0122500, and the Hong Kong Research Grants Council (RGC) General Research Fund (GRF 14204524, 14206125). E.W., Ru.J., Z.L., N.Z., Xi.Z., and H.R. received funding from NSFC/RGC Joint Research Scheme (N\_CUHK420/22). X.G. and H.C. were supported by the Hong Kong Research Grants Council Early Career Scheme grant 22203525. Mat.W., Pr.K., Sab.M., and M.N. received funding from the European Union’s Horizon 2020 research and agreement No 857533. The contribution was made within the project of the Minister of Science and Higher Education "Support for the activity of Centers of Excellence established in Poland under Horizon 2020" on the basis of the contract number MEiN/2023/DIR/3796. Ra.Y., O.H., Mar.W., S.B., and S.S. were supported by the German Research Foundation (DFG, Deutsche Forschungsgemeinschaft) as part of Germany’s Excellence Strategy - EXC 2050/1 - Project ID 390696704 - Cluster of Excellence “Centre for Tactile Internet with Human-in-the-Loop” (CeTI). A.R.J. was supported by the Federal Ministry of Research, Technology, and Space (BMFTR) as part of the research program Communication Systems “Souver\"an. Digital. Vernetzt.”. Joint project 6G-life, project identification number: 16KIS2413K. Lo.M. was supported by the project “Next Generation AI Computing (gAIn),” funded by the Bavarian Ministry of Science and the Arts and the Saxon Ministry for Science, Culture, and Tourism. F.A., M.R.J., Si.K., and So.K. were supported by the National Cancer Institute of the National Institutes of Health under Award Number R21CA280747, the IC2 Institute, and Texas Robotics Seed Collaborative Funding at the University of Texas at Austin. Al.A., Giu.D., Fe.B., Ke.H., Gio.D., Fed.L., Lu.M., and C.A.A. were supported by the European Research Council (ERC) under the Horizon Europe programme (Grant EndoTheranostics, Grant Agreement No. 101118626, https://doi.org/10.3030/101118626), Funded by the European Union. Views and opinions expressed are, however, those of the author(s) only and do not necessarily reflect those of the European Union, the European Research Council Executive Agency, or the Health and Digital Executive Agency. Neither the European Union nor the granting authority can be held responsible for them.

\paragraph*{Author contributions:}
Data Contribution: Everyone,
Conceptualization: N.N., J.C., D.H., Z.J., A.K., S.D.H., M.A.;
Methodology:  N.N., J.C., A.K., S.D.H., L.Z.;
Evaluation:  N.N., J.C., Je.H., Xin.C., D.G.M., P.T.;
Software:  N.N., J.C., L.Z.;
Visualization: N.N., J.C., Je.H., L.Z.;
Data Curation:  N.N., J.C., Xin.C., D.H.;
Formal analysis: N.N., Je.H.;
Funding acquisition: A.K., S.D.H., M.A., Na.N.;
Supervision: A.K., S.D.H., M.A., Na.N.;
Writing - original draft: Je.H., N.N., J.C., Xin.C., L.Z.;
Writing - review and editing: Je.H., N.N., J.C., Xin.C., L.Z., A.K., S.D.H., Al.K., A.M.O., K.G., T.H., Pe.K.;

\paragraph*{Competing interests:}
The following authors declare competing interests. All other authors have no competing interests to declare.
\begin{itemize}
  \item A.K. is a co-founder of Semaphor Surgical.
  \item F.F. receives consultant fees from Boston Scientific and Johnson \& Johnson. F.F. has received research grants from Intuitive Surgical.
  \item A.M.O. has received research grants from Intuitive Surgical.
\end{itemize}

\paragraph*{Data and materials availability:}
The Open-H-Embodiment dataset is publicly available on Hugging Face under a CC-BY-4.0 license at \url{https://huggingface.co/datasets/nvidia/PhysicalAI-Robotics-Open-H-Embodiment}.
Data collection guidelines, conversion scripts, and validation tools are maintained in the Open-H data collection repository at \url{https://github.com/open-h/open-h-embodiment}.
GR00T-H model weights are available at \url{https://huggingface.co/nvidia/GR00T-H}, with training and inference code at \url{https://github.com/NVIDIA-Medtech/GR00T-H}.
LingBot-VA-Suture model weights are available at \url{https://huggingface.co/semaphorsurgical/lingbot-va-suture}, with code at \url{https://github.com/semaphor-surgical/lingbot-va-suture}.
Cosmos-H-Surgical-Simulator model weights are available at \url{https://huggingface.co/nvidia/Cosmos-H-Surgical-Simulator}, with training, inference, and evaluation code at \url{https://github.com/NVIDIA-Medtech/Cosmos-H-Surgical-Simulator}.
All materials needed to reproduce the results in this paper are accessible through these repositories without restriction beyond the stated licenses.


\subsection*{Supplementary materials}
Supplementary Text\\
Figure S1\\
Tables S1 to S4\\
Movies S1 to S3


\newpage


\renewcommand{\thefigure}{S\arabic{figure}}
\renewcommand{\thetable}{S\arabic{table}}
\renewcommand{\theequation}{S\arabic{equation}}
\renewcommand{\thepage}{S\arabic{page}}
\setcounter{figure}{0}
\setcounter{table}{0}
\setcounter{equation}{0}
\setcounter{page}{1}


\begin{center}
\section*{Supplementary Materials for\\ \scititle}

Open-H-Embodiment Consortium\\
Nigel Nelson, et al.
\end{center}

\subsubsection*{This PDF file includes:}
Supplementary Text\\
Figure S1\\
Tables S1 to S4\\
Movies S1 to S3

\newpage


\subsection*{Supplementary Text}

\subsubsection*{World Action Model Baseline: LingBot-VA}
\label{sec:lingbot_va_suture}

To assess whether video-based world-action modeling offers a viable alternative for accurate action prediction in surgical autonomy, we adapt LingBot-VA~\cite{li2026causal} to the SutureBot suturing task, producing LingBot-VA-Suture. LingBot-VA-Suture is initialized from the Wan2.2-5B video diffusion model: at each autoregressive step it predicts a chunk of future latent visual states via conditional flow matching and simultaneously decodes the corresponding end-effector actions through inverse dynamics. This formulation decouples visual dynamics modeling from action control, in contrast to GR00T-H and ACT, which couple both within a single observation-to-action mapping.

\paragraph*{Surgical Adaptation.}
We follow a two-stage procedure. The base LingBot-VA checkpoint is first further pretrained on 50 hours of JHU IMERSE dVRK-Si datasets from Open-H-Embodiment to acquire surgical visual and kinematic priors, and then post-trained on the SutureBot dataset at a reduced learning rate. LingBot-VA-Suture processes three synchronized camera views (endoscope, left wrist, and right wrist) and outputs 16-dimensional actions (7-DoF end-effector pose plus gripper jaw angle per arm) at a chunk size of 48 actions at 30\,Hz with quantile-based action normalization, together with video predictions at 10\,Hz.
\newline


\begin{figure}
    \centering
    \includegraphics[width=1.0\linewidth]{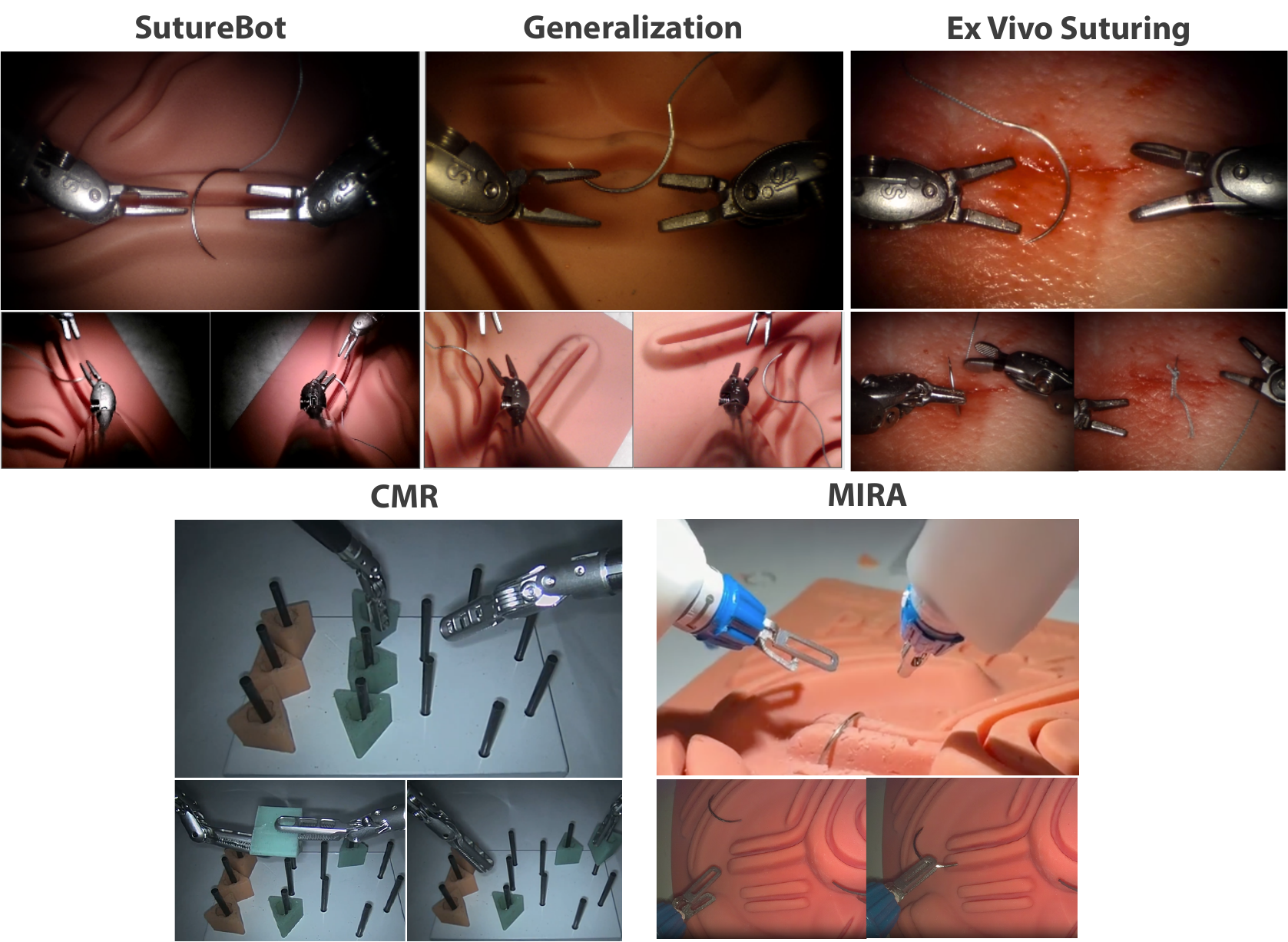}
    \caption{\textbf{Experimental setups for GR00T-H policy evaluation.}
    SutureBot: The primary evaluation environment, utilizing a da Vinci Research Kit Si (dVRK-Si), silicone phantom pad, and suture needle. Generalization: An out-of-distribution testbed featuring wound geometries absent from the training data and external lighting sources to replace standard endoscope illumination. MIRA and CMR: Robotic platforms used to validate multi-embodiment policy performance. Ex vivo suturing: Performed with skin-on pork belly to assess performance on real tissue.}
    \label{fig:eval_setups}
\end{figure}

\begin{figure}
    \captionsetup{labelformat=empty}
    \centering
    \includegraphics[width=1.0\linewidth]{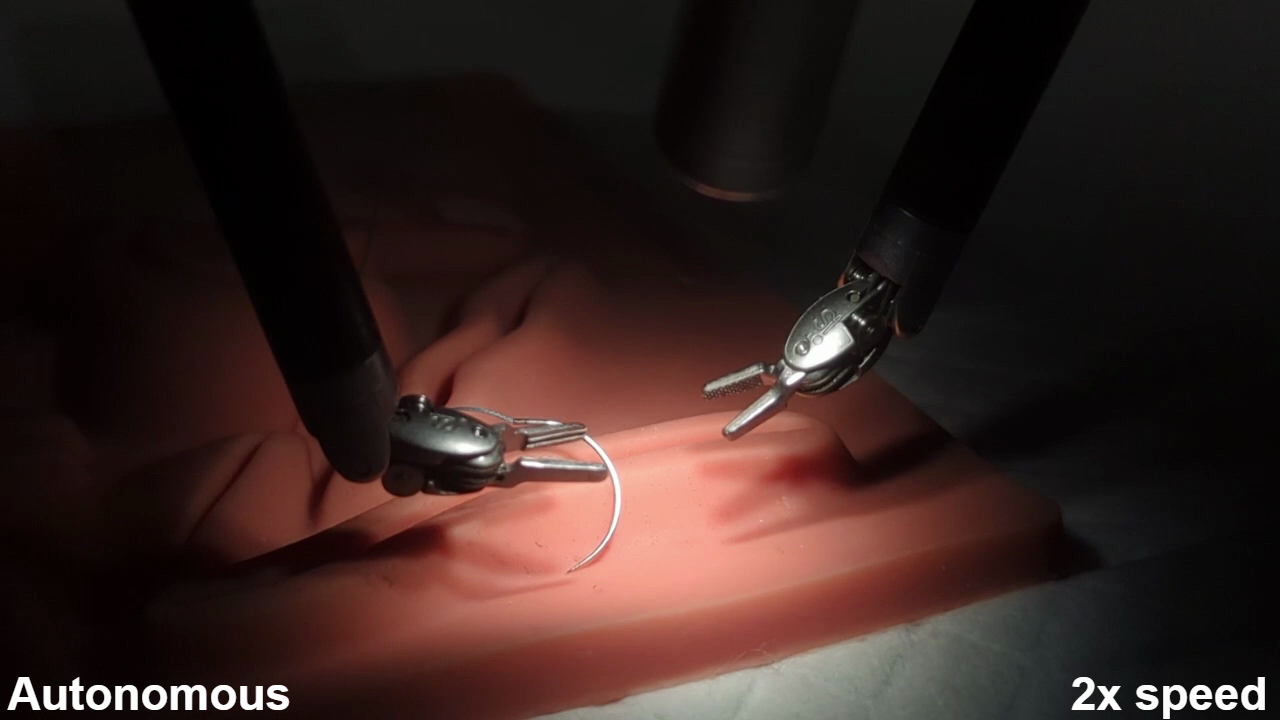}
    \caption{\textbf{\href{\groothendtoendurl}{Movie S1}:} Successful end-to-end suturing rollout by GR00T-H on the SutureBot benchmark. GR00T-H controls the dVRK-Si to pick up and hand over the needle, perform needle throw and extraction, and complete the task with a knot tie.}
    \label{fig:gr00t_h_e2e}
\end{figure}

\begin{figure}
    \captionsetup{labelformat=empty}
    \centering
    \includegraphics[width=1.0\linewidth]{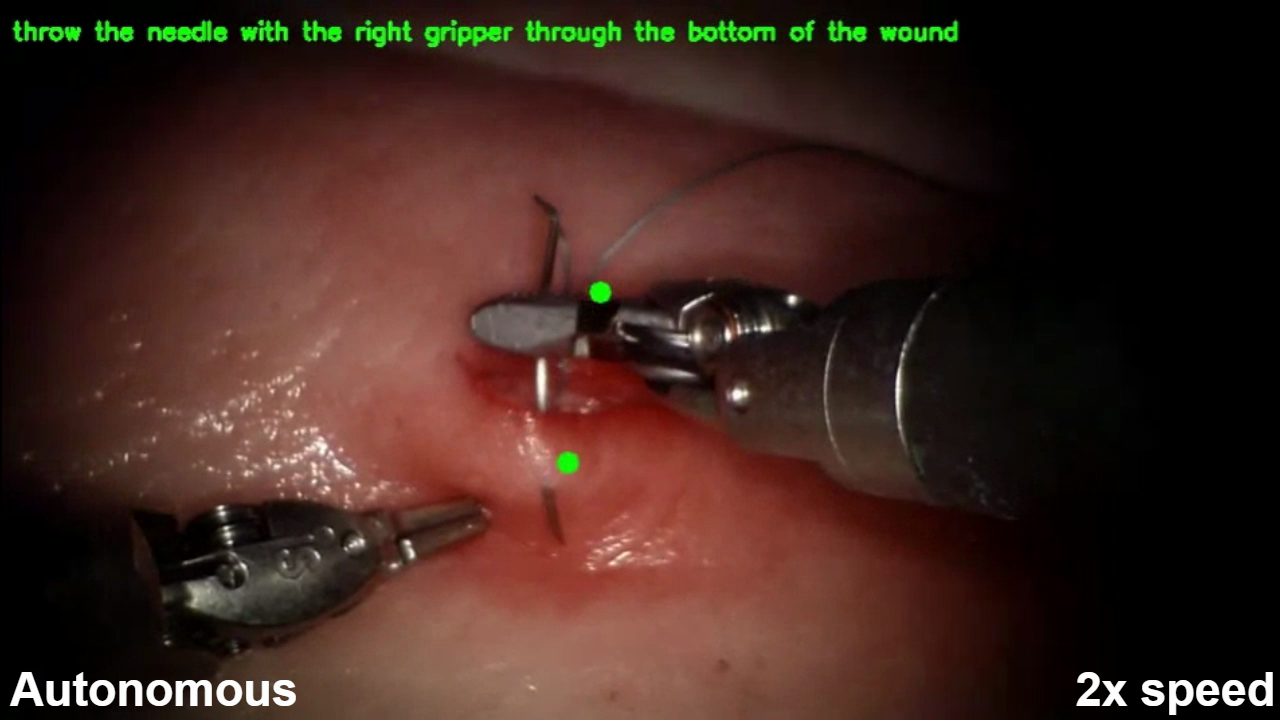}
    \caption{\textbf{\href{\groothwoundclosureurl}{Movie S2}:} Successful sub-task rollouts by GR00T-H for ex vivo suturing. The policy is conditioned on current observations and sub-task prompts, with the prompts shown in the top left of the video.}
    \label{fig:gr00t_h_wound_closure}
\end{figure}

\begin{figure}
    \captionsetup{labelformat=empty}
    \centering
    \includegraphics[width=1.0\linewidth]{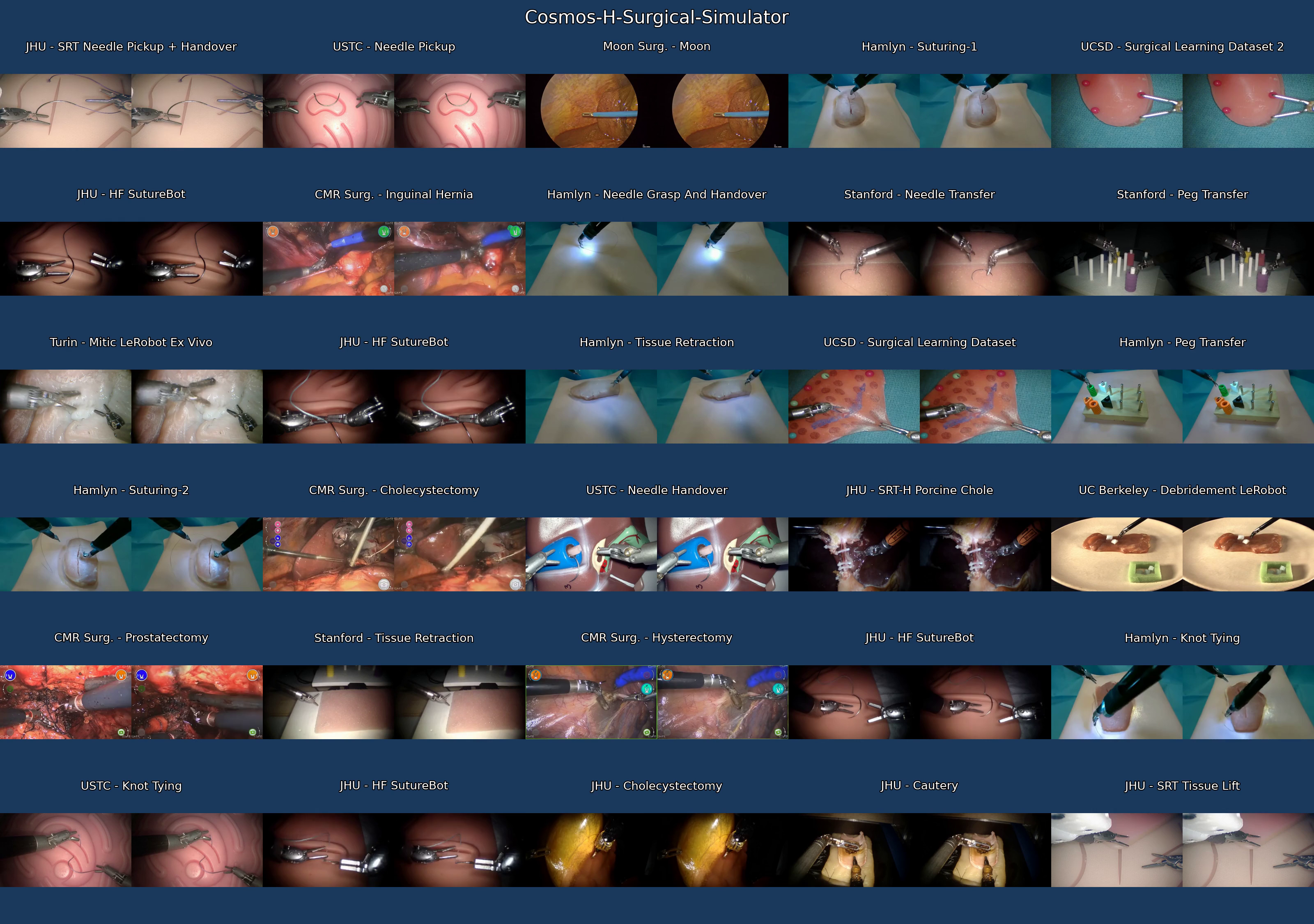}
    \caption{\textbf{\href{\chssvideourl}{Movie S3}:} Qualitative results from Cosmos-H-Surgical-Simulator across 30 Open-H datasets, 9 institutions, and 9 embodiments. Each panel shows ground-truth observations (left) alongside model-predicted frames (right), conditioned on recorded kinematic action trajectories.}
    \label{fig:c_h_s_s_viz}
\end{figure}


\begingroup
\renewcommand{\arraystretch}{0.82}
\setlength{\tabcolsep}{2pt}
\hbadness=10000
\tiny
\begin{longtable}{>{\raggedright\arraybackslash}p{1.4cm} >{\raggedright\arraybackslash}p{2.8cm} >{\raggedright\arraybackslash}p{3.8cm} >{\raggedright\arraybackslash}p{1.5cm} >{\raggedright\arraybackslash}p{2.2cm} >{\raggedright\arraybackslash}p{1.8cm}}
\caption{\textbf{Complete Open-H-Embodiment dataset inventory.} Each row corresponds to one contributed dataset in the final release. The Scale column lists episodes / frames / hours. A dagger ($^{\dagger}$) after the dataset name indicates that ground-truth kinematics are not included in the released dataset. Abbreviations: Ster.\ = stereo endoscope, Endo.\ = endoscope, US = ultrasound, Wr.\ = wrist camera, D = depth, TPV = third-person view, Seg.\ = segmentation, HO = handover, PU = pickup, Sim.\ = simulated/simulation, Bench = benchtop/phantom, Clin.\ = clinical, Ex V.\ = ex vivo, In V.\ = in vivo.}
\label{tab:dataset_inventory} \\
\hline
\textbf{Group} & \textbf{Dataset} & \textbf{Domain / Task [Env.]} & \textbf{Scale} & \textbf{Robot} & \textbf{Views} \\
\hline
\endfirsthead
\multicolumn{6}{l}{\textit{Table~\ref{tab:dataset_inventory} continued}} \\
\hline
\textbf{Group} & \textbf{Dataset} & \textbf{Domain / Task [Env.]} & \textbf{Scale} & \textbf{Robot} & \textbf{Views} \\
\hline
\endhead
\hline
\multicolumn{6}{r}{\textit{Continued on next page}} \\
\endfoot
\hline
\endlastfoot
Balgrist & sonogym\_\allowbreak{}open\_\allowbreak{}h\_\allowbreak{}US\_\allowbreak{}guidance\_\allowbreak{}L1 & US -- Robotic US (US Guidance) [Sim.] & 1,024 ep\newline 183,296 fr\newline 0.42 h & Sim. KUKA Med14 & US \\
Balgrist & sonogym\_\allowbreak{}open\_\allowbreak{}h\_\allowbreak{}US\_\allowbreak{}guidance\_\allowbreak{}L2 & US -- Robotic US (US Guidance) [Sim.] & 1,024 ep\newline 183,296 fr\newline 0.42 h & Sim. KUKA Med14 & US \\
Balgrist & sonogym\_\allowbreak{}open\_\allowbreak{}h\_\allowbreak{}US\_\allowbreak{}guidance\_\allowbreak{}L3 & US -- Robotic US (US Guidance) [Sim.] & 1,024 ep\newline 183,296 fr\newline 0.42 h & Sim. KUKA Med14 & US \\
Balgrist & sonogym\_\allowbreak{}open\_\allowbreak{}h\_\allowbreak{}US\_\allowbreak{}guidance\_\allowbreak{}L4 & US -- Robotic US (US Guidance) [Sim.] & 1,024 ep\newline 183,296 fr\newline 0.42 h & Sim. KUKA Med14 & US \\
Balgrist & sonogym\_\allowbreak{}open\_\allowbreak{}h\_\allowbreak{}US\_\allowbreak{}guidance\_\allowbreak{}L5 & US -- Robotic US (US Guidance) [Sim.] & 1,024 ep\newline 183,296 fr\newline 0.42 h & Sim. KUKA Med14 & US \\
Balgrist & ultrabones\_\allowbreak{}lerobot\_\allowbreak{}dataset\_\allowbreak{}full & US -- Robotic US (US Scan) [Bench] & 60 ep\newline 90,371 fr\newline 1.20 h & Sim. KUKA Med14 & US \\
Balgrist & ultrabones\_\allowbreak{}lerobot\_\allowbreak{}dataset\_\allowbreak{}full\_\allowbreak{}2 & US -- Robotic US (US Scan) [Bench] & 12 ep\newline 20,668 fr\newline 0.27 h & Sim. KUKA Med14 & US \\
Balgrist & ultrabones\_\allowbreak{}lerobot\_\allowbreak{}dataset\_\allowbreak{}full\_\allowbreak{}2\_\allowbreak{}synthetic\_\allowbreak{}robot\_\allowbreak{}2 & US -- Robotic US (US Scan) [Sim.] & 12 ep\newline 20,668 fr\newline 0.27 h & Sim. KUKA Med14 & US \\
Balgrist & ultrabones\_\allowbreak{}lerobot\_\allowbreak{}dataset\_\allowbreak{}full\_\allowbreak{}3 & US -- Robotic US (US Scan) [Bench] & 18 ep\newline 24,104 fr\newline 0.32 h & Sim. KUKA Med14 & US \\
Balgrist & ultrabones\_\allowbreak{}lerobot\_\allowbreak{}dataset\_\allowbreak{}full\_\allowbreak{}3\_\allowbreak{}synthetic\_\allowbreak{}robot\_\allowbreak{}2 & US -- Robotic US (US Scan) [Sim.] & 18 ep\newline 24,104 fr\newline 0.32 h & Sim. KUKA Med14 & US \\
Balgrist & ultrabones\_\allowbreak{}lerobot\_\allowbreak{}dataset\_\allowbreak{}full\_\allowbreak{}synthetic\_\allowbreak{}robot & US -- Robotic US (US Scan) [Sim.] & 60 ep\newline 90,371 fr\newline 1.20 h & Sim. KUKA Med14 & US \\
\hline
CMR Surg. & cholecystectomy & Surg. -- Cholecystectomy (Chole.) [Clin.] & 4,792 ep\newline 16,999,777 fr\newline 78.70 h & Versius & Endo. \\
CMR Surg. & drybox & Surg. -- Skills Benchmark (Peg Transfer) [Bench] & 480 ep\newline 1,677,540 fr\newline 7.77 h & Versius & Endo. \\
CMR Surg. & hysterectomy & Surg. -- Hysterectomy (Hysterect.) [Clin.] & 7,432 ep\newline 26,374,851 fr\newline 122.11 h & Versius & Endo. \\
CMR Surg. & inguinal\_\allowbreak{}hernia & Surg. -- Hernia Repair (Hernia Rep.) [Clin.] & 7,286 ep\newline 25,807,467 fr\newline 119.48 h & Versius & Endo. \\
CMR Surg. & peg\_\allowbreak{}transfer\_\allowbreak{}extra & Surg. -- Skills Benchmark (Peg Transfer) [Bench] & 216 ep\newline 762,828 fr\newline 3.53 h & Versius & Endo. \\
CMR Surg. & prostatectomy & Surg. -- Prostatectomy (Prostatect.) [Clin.] & 10,329 ep\newline 36,516,224 fr\newline 169.06 h & Versius & Endo. \\
\hline
CUHK & bmt\_\allowbreak{}insertion\_\allowbreak{}dataset & US -- US-Guided Interv. (Needle Ins.) [Ex V.] & 847 ep\newline 814,451 fr\newline 7.54 h & UR5e & US,\newline TPV \\
CUHK & bmt\_\allowbreak{}insertion\_\allowbreak{}tumor\_\allowbreak{}fishball\_\allowbreak{}dataset & US -- US-Guided Interv. (Needle Ins.) [Ex V.] & 414 ep\newline 401,977 fr\newline 3.72 h & UR5e & US,\newline TPV \\
CUHK & bmt\_\allowbreak{}insertion\_\allowbreak{}vessel\_\allowbreak{}dataset & US -- US-Guided Interv. (Needle Ins.) [Ex V.] & 665 ep\newline 637,649 fr\newline 5.90 h & UR5e & US,\newline TPV \\
CUHK & bmt\_\allowbreak{}needle\_\allowbreak{}insertion\_\allowbreak{}dataset & US -- US-Guided Interv. (Needle Ins.) [Ex V.] & 386 ep\newline 376,835 fr\newline 3.49 h & UR5e & US,\newline TPV \\
CUHK & bmt\_\allowbreak{}tumor(grape)\_\allowbreak{}insertion\_\allowbreak{}dataset & US -- US-Guided Interv. (Needle Ins.) [Ex V.] & 99 ep\newline 107,054 fr\newline 0.99 h & UR5e & US,\newline TPV \\
CUHK & US\_\allowbreak{}PPNR & US -- Probe Placement and Needle Retrieval [Ex V.] & 2,546 ep\newline 1,172,070 fr\newline 10.85 h & UR5e & US,\newline Wr.,\newline D,\newline TPV \\
CUHK & Find\_\allowbreak{}greater\_\allowbreak{}curvature & Endo. -- Flex. Endoscopy (Anat. Loc.) [Bench] & 462 ep\newline 107,488 fr\newline 1.49 h & Custom Endo. Rob. & Endo. \\
CUHK & Find\_\allowbreak{}lesser\_\allowbreak{}curvature & Endo. -- Flex. Endoscopy (Anat. Loc.) [Bench] & 200 ep\newline 61,997 fr\newline 0.86 h & Custom Endo. Rob. & Endo. \\
CUHK & Find\_\allowbreak{}pyloric\_\allowbreak{}antrum & Endo. -- Flex. Endoscopy (Anat. Loc.) [Bench] & 204 ep\newline 28,731 fr\newline 0.40 h & Custom Endo. Rob. & Endo. \\
CUHK & Track\_\allowbreak{}orange\_\allowbreak{}lesion & Endo. -- Flex. Endoscopy (Lesion Track.) [Bench] & 156 ep\newline 33,345 fr\newline 0.46 h & Custom Endo. Rob. & Endo. \\
CUHK & Track\_\allowbreak{}small\_\allowbreak{}round\_\allowbreak{}nodule & Endo. -- Flex. Endoscopy (Lesion Track.) [Bench] & 722 ep\newline 249,461 fr\newline 3.46 h & Custom Endo. Rob. & Endo. \\
CUHK & Track\_\allowbreak{}white\_\allowbreak{}oval\_\allowbreak{}ROI & Endo. -- Flex. Endoscopy (ROI Track.) [Bench] & 414 ep\newline 98,908 fr\newline 1.37 h & Custom Endo. Rob. & Endo. \\
CUHK & Tracked\_\allowbreak{}EUS & US -- Endo. US (Tracked EUS) [In V.] & 34 ep\newline 33,425 fr\newline 0.37 h & dVRK & Endo.,\newline US \\
CUHK & Tracked\_\allowbreak{}US & US -- Robotic US (Tracked US) [Clin.] & 30 ep\newline 7,007 fr\newline 0.08 h & dVRK & US,\newline TPV \\
\hline
Hamlyn/ Imperial & knot\_\allowbreak{}tying & Surg. -- Suture/Knot (Knot Tying) [Ex V.] & 49 ep\newline 30,222 fr\newline 0.28 h & dVRK & Endo.,\newline Wr.,\newline D \\
Hamlyn/ Imperial & needle\_\allowbreak{}grasp\_\allowbreak{}and\_\allowbreak{}handover & Surg. -- Suture/Knot (Needle HO) [Bench] & 137 ep\newline 46,582 fr\newline 0.43 h & dVRK & Endo.,\newline Wr.,\newline D \\
Hamlyn/ Imperial & peg\_\allowbreak{}transfer & Surg. -- Skills Benchmark (Peg Transfer) [Bench] & 317 ep\newline 102,187 fr\newline 0.95 h & dVRK & Endo.,\newline Wr.,\newline D \\
Hamlyn/ Imperial & Suturing-1 & Surg. -- Suture/Knot (Suturing) [Ex V.] & 180 ep\newline 100,067 fr\newline 0.93 h & dVRK & Endo.,\newline Wr.,\newline D \\
Hamlyn/ Imperial & Suturing-2 & Surg. -- Suture/Knot (Suturing) [Ex V.] & 186 ep\newline 251,355 fr\newline 2.33 h & dVRK & Endo.,\newline Wr.,\newline D \\
Hamlyn/ Imperial & tissue\_\allowbreak{}lifting & Surg. -- Tissue Manip. (Lifting) [Ex V.] & 75 ep\newline 8,180 fr\newline 0.15 h & dVRK & Endo.,\newline Wr.,\newline D \\
Hamlyn/ Imperial & Tissue\_\allowbreak{}Retraction & Surg. -- Tissue Manip. (Retraction) [Ex V.] & 75 ep\newline 14,160 fr\newline 0.13 h & dVRK & Endo.,\newline Wr.,\newline D \\
\hline
HKBU & 46\_\allowbreak{}datasets\_\allowbreak{}summary & US -- Robotic US (US Nav.) [Sim.] & 6,072 ep\newline 322,000 fr\newline 4.47 h & Sim. US Platform & US \\
\hline
ImFusion & ImFusion\_\allowbreak{}dataset\_\allowbreak{}corrected & US -- Robotic US (US Scan) [Bench] & 1,644 ep\newline 559,897 fr\newline 5.18 h & Franka Panda & US,\newline Wr.,\newline TPV \\
\hline
JHU & cao\_\allowbreak{}cautery\_\allowbreak{}combined & Surg. -- Tissue Manip. (Debride.) [Ex V.] & 783 ep\newline 52,748 fr\newline 0.49 h & dVRK-Si & Ster.,\newline Wr. \\
JHU & cautery & Surg. -- Tissue Manip. (Debride.) [Ex V.] & 22 ep\newline 5,288 fr\newline 0.05 h & dVRK-Si & Ster.,\newline Wr. \\
JHU & Cholecystectomy & Surg. -- Cholecystectomy (Chole.) [Ex V.] & 750 ep\newline 181,021 fr\newline 1.68 h & dVRK-Si & Ster.,\newline Wr. \\
JHU & endosrt & Endo. -- Flex. Endoscopy (Endo. Nav.) [Bench] & 2,047 ep\newline 193,203 fr\newline 2.15 h & Flex. Endo. Platf. & Other \\
JHU & hf\_\allowbreak{}suturebot & Surg. -- Suture/Knot (Suturing) [Bench] & 1,452 ep\newline 516,334 fr\newline 4.78 h & dVRK-Si & Ster.,\newline Wr. \\
JHU & nephfat & Surg. -- Tissue Manip. (Debride.) [Bench] & 2,112 ep\newline 494,525 fr\newline 4.58 h & dVRK-Si & Ster.,\newline Wr. \\
JHU & SurgSync-stitch-coldcut-P1 & Surg. -- Suture/Knot (Suturing) [Ex V.] & 578 ep\newline 53,114 fr\newline 1.48 h & dVRK-Si & Endo. \\
JHU & SurgSync-stitch-coldcut-P2 & Surg. -- Suture/Knot (Suturing) [Ex V.] & 313 ep\newline 32,426 fr\newline 0.90 h & dVRK-Si & Endo. \\
JHU & SurgSync-stitch-coldcut-P3 & Surg. -- Suture/Knot (Suturing) [Ex V.] & 196 ep\newline 17,485 fr\newline 0.49 h & dVRK-Si & Endo. \\
JHU & Prepare to Pierce & Surg. -- Suture/Knot (Needle Pos.) [Bench] & 2 ep\newline 582 fr\newline 0.01 h & dVRK-Si & Ster. \\
JHU & srt\_\allowbreak{}needle\_\allowbreak{}pickup+handover & Surg. -- Suture/Knot (Needle PU) [Bench] & 240 ep\newline 58,305 fr\newline 0.54 h & dVRK & Ster.,\newline Wr. \\
JHU & srt\_\allowbreak{}tissue\_\allowbreak{}lift & Surg. -- Tissue Manip. (Lifting) [Bench] & 225 ep\newline 27,487 fr\newline 0.25 h & dVRK & Ster.,\newline Wr. \\
JHU & srth\_\allowbreak{}porcine\_\allowbreak{}chole\_\allowbreak{}fix & Surg. -- Cholecystectomy (Chole.) [Ex V.] & 14,836 ep\newline 1,878,393 fr\newline 17.39 h & dVRK-Si & Ster.,\newline Wr. \\
JHU & star\_\allowbreak{}IL & Surg. -- Suture/Knot (Suturing) [Ex V.] & 512 ep\newline 216,140 fr\newline 2.00 h & KUKA LBR iiwa & Endo.,\newline Wr. \\
JHU & wound\_\allowbreak{}closure & Surg. -- Suture/Knot (Suturing) [Bench] & 216 ep\newline 1,883,971 fr\newline 17.44 h & dVRK-Si & Ster.,\newline Wr. \\
\hline
Moon Surg. & moon & Surg. -- Camera/View Mgmt. (Cam. Guidance) [Clin.] & 65 ep\newline 12,020 fr\newline 0.11 h & Maestro & Endo.,\newline TPV \\
\hline
Obuda & FRS\_\allowbreak{}Dome\_\allowbreak{}1 & Surg. -- Suture/Knot (Suturing) [Bench] & 102 ep\newline 141,078 fr\newline 1.31 h & dVRK & Ster.,\newline Wr.,\newline D \\
Obuda & NeedleThreading\_\allowbreak{}1 & Surg. -- Suture/Knot (Suturing) [Bench] & 196 ep\newline 103,067 fr\newline 0.95 h & dVRK & Ster.,\newline Wr.,\newline D \\
Obuda & NeedleThreading\_\allowbreak{}2 & Surg. -- Suture/Knot (Suturing) [Bench] & 204 ep\newline 102,221 fr\newline 0.95 h & dVRK & Ster.,\newline Wr.,\newline D \\
Obuda & PegTransfer\_\allowbreak{}1 & Surg. -- Skills Benchmark (Peg Transfer) [Bench] & 216 ep\newline 134,832 fr\newline 1.25 h & dVRK & Ster.,\newline Wr.,\newline D \\
Obuda & PegTransfer\_\allowbreak{}2 & Surg. -- Skills Benchmark (Peg Transfer) [Bench] & 184 ep\newline 78,140 fr\newline 0.72 h & dVRK & Ster.,\newline Wr.,\newline D \\
Obuda & Pork\_\allowbreak{}1 & Surg. -- Tissue Manip. (Grasping) [Ex V.] & 318 ep\newline 165,486 fr\newline 1.53 h & dVRK & Ster.,\newline Wr.,\newline D \\
Obuda & Rollercoaster\_\allowbreak{}1 & Surg. -- Skills Benchmark (Spatial Nav.) [Bench] & 95 ep\newline 130,268 fr\newline 1.21 h & dVRK & Ster.,\newline Wr.,\newline D \\
Obuda & Seaspike\_\allowbreak{}1 & Surg. -- Skills Benchmark (Ring Transfer) [Bench] & 207 ep\newline 89,269 fr\newline 0.83 h & dVRK & Ster.,\newline Wr.,\newline D \\
Obuda & Seaspike\_\allowbreak{}2 & Surg. -- Skills Benchmark (Ring Transfer) [Bench] & 153 ep\newline 67,658 fr\newline 0.63 h & dVRK & Ster.,\newline Wr.,\newline D \\
Obuda & Seaspike\_\allowbreak{}3 & Surg. -- Skills Benchmark (Ring Transfer) [Bench] & 219 ep\newline 102,948 fr\newline 0.95 h & dVRK & Ster.,\newline Wr.,\newline D \\
Obuda & Skinphantom\_\allowbreak{}1 & Surg. -- Suture/Knot (Suturing) [Bench] & 106 ep\newline 41,979 fr\newline 0.39 h & dVRK & Ster.,\newline Wr.,\newline D \\
\hline
HK PolyU & OpenH\_\allowbreak{}Dataset\_\allowbreak{}full & Surg. -- Cholecystectomy (Retraction) [Sim.] & 11,520 ep\newline 5,760,000 fr\newline 53.33 h & Sim. dVRK & Endo. \\
\hline
Rob Surg. & all\_\allowbreak{}merged\_\allowbreak{}data (hemicolectomy) & Surg. -- Clinical Proc. (Hemicolect.) [Clin.] & 5 ep\newline 1,003,887 fr\newline 7.98 h & BiTrack & Endo. \\
Rob Surg. & all\_\allowbreak{}merged\_\allowbreak{}data (hysterectomy) & Surg. -- Clinical Proc. (Hysterect.) [Clin.] & 1 ep\newline 1,003,887 fr\newline 1.32 h & BiTrack & Endo. \\
\hline
SanoScience & Expert\_\allowbreak{}demonstrations & Surg. -- Cholecystectomy (Chole.) [Sim.] & 5,454 ep\newline 603,054 fr\newline 6.70 h & XR Simulator & Endo.,\newline D,\newline Seg. \\
SanoScience & NonExpert\_\allowbreak{}failure & Surg. -- Cholecystectomy (Chole.) [Sim.] & 126 ep\newline 17,622 fr\newline 0.16 h & XR Simulator & Endo.,\newline D,\newline Seg. \\
SanoScience & NonExpert\_\allowbreak{}full\_\allowbreak{}modalities & Surg. -- Cholecystectomy (Chole.) [Sim.] & 6,156 ep\newline 713,070 fr\newline 6.60 h & XR Simulator & Endo.,\newline D,\newline Seg. \\
SanoScience & NonExpert\_\allowbreak{}partial\_\allowbreak{}modalities & Surg. -- Cholecystectomy (Chole.) [Sim.] & 666 ep\newline 58,752 fr\newline 0.54 h & XR Simulator & Endo. \\
SanoScience & NonExpert\_\allowbreak{}recovery & Surg. -- Cholecystectomy (Chole.) [Sim.] & 126 ep\newline 20,376 fr\newline 0.19 h & XR Simulator & Endo.,\newline D,\newline Seg. \\
SanoScience & NonExpert\_\allowbreak{}stereo & Surg. -- Cholecystectomy (Chole.) [Sim.] & 1,602 ep\newline 179,640 fr\newline 1.66 h & XR Simulator & Ster.,\newline D,\newline Seg. \\
\hline
Semaphor & open\_\allowbreak{}h\_\allowbreak{}semaphor & Surg. -- Suture/Knot (Suturing) [Ex V.] & 535 ep\newline 47,050 fr\newline 0.44 h & Manual Lap. Tools & TPV \\
Semaphor & open\_\allowbreak{}h\_\allowbreak{}semaphor\_\allowbreak{}1.18 & Surg. -- Suture/Knot (Suturing) [Ex V.] & 470 ep\newline 50,664 fr\newline 0.47 h & Manual Lap. Tools & TPV \\
\hline
Stanford & block\_\allowbreak{}transfer\_\allowbreak{}sim\_\allowbreak{}lerobot\_\allowbreak{}1\_\allowbreak{}28 & Surg. -- Skills Benchmark (Peg Transfer) [Sim.] & 500 ep\newline 236,968 fr\newline 3.46 h & Sim. dVRK & Ster. \\
Stanford & Needle Transfer & Surg. -- Suture/Knot (Needle HO) [Bench] & 700 ep\newline 313,882 fr\newline 2.91 h & dVRK-Si & Ster. \\
Stanford & needle\_\allowbreak{}transfer\_\allowbreak{}sim\_\allowbreak{}lerobot\_\allowbreak{}1\_\allowbreak{}28 & Surg. -- Suture/Knot (Needle HO) [Sim.] & 500 ep\newline 229,232 fr\newline 3.35 h & Sim. dVRK & Ster. \\
Stanford & Peg Transfer & Surg. -- Skills Benchmark (Peg Transfer) [Bench] & 598 ep\newline 268,729 fr\newline 2.49 h & dVRK-Si & Ster. \\
Stanford & Tissue Retraction & Surg. -- Tissue Manip. (Retraction) [Bench] & 698 ep\newline 291,826 fr\newline 2.70 h & dVRK-Si & Ster. \\
\hline
TU Dresden & endoscope\_\allowbreak{}guidance & Surg. -- Camera/View Mgmt. (Cam. Guidance) [In V.] & 50 ep\newline 58,605 fr\newline 0.54 h & UR5e & Endo. \\
TU Dresden & endoscope\_\allowbreak{}guidance & Surg. -- Camera/View Mgmt. (Cam. Guidance) [In V.] & 50 ep\newline 56,484 fr\newline 0.52 h & UR5e & Endo. \\
TU Dresden & grasping\_\allowbreak{}retraction & Surg. -- Tissue Manip. (Retraction) [In V.] & 116 ep\newline 35,406 fr\newline 0.33 h & UR5e & Endo. \\
TU Dresden & grasping\_\allowbreak{}retraction & Surg. -- Tissue Manip. (Retraction) [In V.] & 146 ep\newline 47,753 fr\newline 0.44 h & UR5e & Endo. \\
\hline
TUM CAMP & SonATA\_\allowbreak{}all & US -- Robotic US (US Scan) [Bench] & 2,397 ep\newline 633,604 fr\newline 5.87 h & Franka Panda & US,\newline Wr.,\newline TPV \\
\hline
Turin & mitic\_\allowbreak{}lerobot\_\allowbreak{}ex\_\allowbreak{}vivo & Surg. -- Suture/Knot (Suturing) [Ex V.] & 799 ep\newline 388,690 fr\newline 3.60 h & dVRK & Ster. \\
Turin & mitic\_\allowbreak{}lerobot\_\allowbreak{}plastic\_\allowbreak{}pad & Surg. -- Suture/Knot (Suturing) [Bench] & 550 ep\newline 149,846 fr\newline 1.39 h & dVRK & Ster. \\
Turin & mitic\_\allowbreak{}lerobot\_\allowbreak{}plastic\_\allowbreak{}pad\_\allowbreak{}3DMED & Surg. -- Suture/Knot (Suturing) [Bench] & 370 ep\newline 243,229 fr\newline 2.25 h & dVRK & Ster. \\
Turin & mitic\_\allowbreak{}lerobot\_\allowbreak{}plastic\_\allowbreak{}tube & Surg. -- Suture/Knot (Suturing) [Bench] & 480 ep\newline 216,070 fr\newline 2.00 h & dVRK & Ster. \\
\hline
UBC & GauzeCutting\_\allowbreak{}merged$^{\dagger}$ & Surg. -- Benchtop Tasks (Gauze Cut.) [Bench] & 234 ep\newline 75,277 fr\newline 0.69 h & da Vinci Si & Ster. \\
UBC & KnotTying\_\allowbreak{}merged$^{\dagger}$ & Surg. -- Suture/Knot (Knot Tying) [Bench] & 523 ep\newline 75,840 fr\newline 0.70 h & da Vinci Si & Ster. \\
UBC & NeedlePassing\_\allowbreak{}merged$^{\dagger}$ & Surg. -- Suture/Knot (Needle HO) [Bench] & 685 ep\newline 70,915 fr\newline 0.67 h & da Vinci Si & Ster. \\
UBC & PegTransfering\_\allowbreak{}merged$^{\dagger}$ & Surg. -- Skills Benchmark (Peg Transfer) [Bench] & 550 ep\newline 14,305 fr\newline 0.13 h & da Vinci Si & Ster. \\
UBC & PickAndPlace\_\allowbreak{}merged$^{\dagger}$ & Surg. -- Tissue Manip. (Grasping) [Bench] & 702 ep\newline 35,681 fr\newline 0.33 h & da Vinci Si & Ster. \\
UBC & Suturing\_\allowbreak{}merged$^{\dagger}$ & Surg. -- Suture/Knot (Suturing) [Bench] & 606 ep\newline 103,917 fr\newline 0.97 h & da Vinci Si & Ster. \\
UBC & WireChasing3D\_\allowbreak{}merged$^{\dagger}$ & Surg. -- Skills Benchmark (Wire Chasing) [Bench] & 523 ep\newline 41,102 fr\newline 0.38 h & da Vinci Si & Ster. \\
UBC & WireChasing\_\allowbreak{}merged$^{\dagger}$ & Surg. -- Skills Benchmark (Wire Chasing) [Bench] & 615 ep\newline 60,042 fr\newline 0.55 h & da Vinci Si & Ster. \\
\hline
UC Berkeley & debridement\_\allowbreak{}lerobot & Surg. -- Tissue Manip. (Debride.) [Ex V.] & 589 ep\newline 221,950 fr\newline 2.06 h & dVRK & Ster. \\
\hline
UCSD & surgical\_\allowbreak{}learning\_\allowbreak{}dataset & Surg. -- Tissue Manip. (Dissection) [Bench] & 912 ep\newline 288,604 fr\newline 2.67 h & dVRK & Ster. \\
UCSD & surgical\_\allowbreak{}learning\_\allowbreak{}dataset2 & Surg. -- Tissue Manip. (Retraction) [Bench] & 200 ep\newline 26,313 fr\newline 0.24 h & dVRK & Ster. \\
UCSD & surgical\_\allowbreak{}learning\_\allowbreak{}retraction\_\allowbreak{}dataset3 & Surg. -- Tissue Manip. (Retraction) [Bench] & 598 ep\newline 183,622 fr\newline 1.70 h & dVRK & Ster. \\
UCSD & surgical\_\allowbreak{}learning\_\allowbreak{}retraction\_\allowbreak{}failurecase & Surg. -- Tissue Manip. (Retraction) [Bench] & 299 ep\newline 66,839 fr\newline 0.62 h & dVRK & Ster. \\
\hline
UIC & UIC\_\allowbreak{}CRCD\_\allowbreak{}LeRobot & Surg. -- Cholecystectomy (Chole.) [Ex V.] & 18 ep\newline 755,891 fr\newline 3.50 h & dVRK & Endo. \\
\hline
USTC/ Tuodao & exvivo\_\allowbreak{}liver\_\allowbreak{}sep & Surg. -- Resect./Dissect. (Liver Dissect.) [Ex V.] & 666 ep\newline 121,922 fr\newline 1.41 h & Torin & Ster. \\
USTC/ Tuodao & grasp\_\allowbreak{}on\_\allowbreak{}liver & Surg. -- Tissue Manip. (Grasping) [Ex V.] & 817 ep\newline 63,538 fr\newline 0.74 h & Torin & Ster. \\
USTC/ Tuodao & invivo\_\allowbreak{}liver\_\allowbreak{}sep & Surg. -- Resect./Dissect. (Liver Dissect.) [In V.] & 199 ep\newline 39,899 fr\newline 0.46 h & Torin & Ster. \\
USTC/ Tuodao & knot\_\allowbreak{}tying & Surg. -- Suture/Knot (Knot Tying) [Bench] & 1,098 ep\newline 182,836 fr\newline 2.12 h & Torin & Ster. \\
USTC/ Tuodao & Needle\_\allowbreak{}handover & Surg. -- Suture/Knot (Needle HO) [Bench] & 260 ep\newline 34,990 fr\newline 0.40 h & Torin & Ster. \\
USTC/ Tuodao & needle\_\allowbreak{}pickup & Surg. -- Suture/Knot (Needle PU) [Bench] & 616 ep\newline 57,172 fr\newline 0.66 h & Torin & Ster. \\
USTC/ Tuodao & tissue\_\allowbreak{}lifting & Surg. -- Tissue Manip. (Lifting) [Bench] & 110 ep\newline 11,673 fr\newline 0.14 h & Torin & Ster. \\
\hline
UT Austin & colonoscope-lerobot & Endo. -- Colonoscopy (Colon. Nav.) [Bench] & 1,894 ep\newline 2,095,587 fr\newline 19.40 h & Cobra Colonoscope & Endo. \\
\hline
UTenn & benchtop\_\allowbreak{}datasets\_\allowbreak{}round2\_\allowbreak{}with\_\allowbreak{}part\_\allowbreak{}seg$^{\dagger}$ & Surg. -- Benchtop Tasks (Tool Track.) [Bench] & 2 ep\newline 223 fr\newline 0.01 h & dVRK & Endo.,\newline D \\
UTenn & surgical\_\allowbreak{}video\_\allowbreak{}datasets$^{\dagger}$ & Surg. -- Surg. Video (Annot.) [Clin.] & 73 ep\newline 8,183 fr\newline 0.08 h & da Vinci Xi & Endo.,\newline D,\newline Seg. \\
UTenn & surgical\_\allowbreak{}video\_\allowbreak{}datasets\_\allowbreak{}round2\_\allowbreak{}with\_\allowbreak{}part\_\allowbreak{}seg$^{\dagger}$ & Surg. -- Surg. Video (Annot.) [Clin.] & 101 ep\newline 8,760 fr\newline 0.08 h & da Vinci Xi & Endo.,\newline D,\newline Seg. \\
UTenn & surgical\_\allowbreak{}video\_\allowbreak{}datasets\_\allowbreak{}with\_\allowbreak{}tip\_\allowbreak{}pose$^{\dagger}$ & Surg. -- Surg. Video (Annot.) [Clin.] & 62 ep\newline 6,690 fr\newline 0.06 h & da Vinci Xi & Endo.,\newline D \\
\hline
Virtual Inc. & Mira Needle Pickup & Surg. -- Benchtop Tasks (Needle PU) [Bench] & 150 ep\newline 39,960 fr\newline 0.37 h & Mira & Endo. \\
\end{longtable}
\endgroup

\begin{table}[t]
\centering
\caption{GR00T-H training parameters.}
\label{tab:gr00t_h_mid_training_details}
\begin{tabular}{ll}
\hline
Parameter & Value \\
\hline
Initialization & \texttt{nvidia/GR00T-N1.6-3B} \\
Frozen backbone & false \\
Action horizon & 50 \\
Action normalization & Temporal z-score \\
Image size & \texttt{224 x 392} \\
Random crop & 95\% \\
Random rotation & 5 deg \\
Coarse pixel dropout & 0.5 \\
Color jitter & Brightness 0.12, contrast 0.15, saturation 0.15, hue 0.02 \\
Multi-view dropout & 0.25 \\
State dropout & 100\% \\
Optimizer & AdamW \\
Learning rate & \texttt{3e-5} \\
Weight decay & \texttt{1e-5} \\
LR schedule & Cosine decay \\
Warmup & 5\% \\
Gradient clipping & 1.0 \\
Global batch size & 1024 \\
Training steps & 65,000 \\
Training compute & 32 A100 80GB x 1.5 days \\
\hline
\end{tabular}
\end{table}

\begingroup
\footnotesize
\renewcommand{\arraystretch}{0.85}
\setlength{\tabcolsep}{4pt}
\begin{longtable}{p{0.72\textwidth}c}
\caption{Open-H dataset mixture used in GR00T-H training} \\
\label{tab:gr00t-h-dataset-mixture} \\
\hline
Dataset / Embodiment Group & Mixture ratio \\
\hline
\endfirsthead

\hline
Dataset / Embodiment Group & Mixture ratio \\
\hline
\endhead

\hline
\endfoot

\hline
\endlastfoot

\textbf{CMR Versius} & \textbf{0.1920} \\
\quad Cholecystectomy & \\
\quad Hysterectomy & \\
\quad Inguinal hernia & \\
\quad Prostatectomy & \\
\quad Drybox peg transfer & \\

\textbf{JHU dVRK-Si} & \textbf{0.3498} \\
\quad ARCADE Cholecystectomy & \\
\quad ARCADE Cautery & \\
\quad IMERSE porcine cholecystectomy & \\
\quad IMERSE CAO cautery & \\
\quad IMERSE needle pickup and handover & \\
\quad IMERSE tissue lift & \\
\quad IMERSE needle pickup & \\
\quad IMERSE wound closure & \\
\quad IMERSE SutureBot & \\
\quad IMERSE NephFat & \\

\textbf{JHU IMERSE monocular} & \textbf{0.0372} \\
\quad SutureBot & \\

\textbf{JHU SMARTS} & \textbf{0.0074} \\
\quad SurgSync stitch coldcut P1 & \\
\quad SurgSync stitch coldcut P2 & \\
\quad SurgSync stitch coldcut P3 & \\

\textbf{Stanford dVRK} & \textbf{0.0630} \\
\quad Needle transfer & \\
\quad Tissue retraction & \\
\quad Peg transfer & \\

\textbf{Obuda dVRK} & \textbf{0.0834} \\
\quad FRS Dome 1 & \\
\quad NeedleThreading 1 & \\
\quad PegTransfer 1 & \\
\quad Rollercoaster 1 & \\
\quad Seaspike 1 & \\
\quad NeedleThreading 2 & \\
\quad PegTransfer 2 & \\
\quad Pork 1 & \\
\quad Seaspike 2 & \\
\quad Seaspike 3 & \\
\quad Skinphantom 1 & \\

\textbf{Rob Surgical} & \textbf{0.0724} \\
\quad All merged data & \\

\textbf{JHU KUKA IMERSE} & \textbf{0.0085} \\
\quad star\_IL & \\

\textbf{USTC / Torin} & \textbf{0.0369} \\
\quad Ex vivo liver separation & \\
\quad Grasp on liver & \\
\quad In vivo liver separation & \\
\quad Knot tying & \\
\quad Needle handover & \\
\quad Needle pickup & \\
\quad Tissue lifting & \\

\textbf{Hamlyn dVRK} & \textbf{0.0393} \\
\quad Suturing-2 & \\
\quad Peg transfer & \\
\quad Suturing-1 & \\
\quad Needle grasp and handover & \\
\quad Knot tying & \\
\quad Tissue retraction & \\

\textbf{UCSD surgical learning} & \textbf{0.0227} \\
\quad Surgical learning dataset & \\
\quad Surgical learning dataset 2 & \\

\textbf{UC Berkeley dVRK} & \textbf{0.0160} \\
\quad Debridement & \\

\textbf{Turin MITIC} & \textbf{0.0680} \\
\quad MITIC ex vivo & \\
\quad Plastic pad & \\
\quad Plastic pad 3DMED & \\
\quad Plastic tube & \\

\textbf{TUD TUNDRA} & \textbf{0.0034} \\
\quad Grasping retraction & \\
\hline
\end{longtable}
\endgroup

\begingroup
\footnotesize
\renewcommand{\arraystretch}{0.85}
\setlength{\tabcolsep}{4pt}
\begin{longtable}{p{0.72\textwidth}c}
\caption{Open-H dataset mixture used in C-H-S-S training} \\
\label{tab:chss-dataset-mixture} \\
\hline
Dataset / Embodiment Group & Mixture ratio \\
\hline
\endfirsthead

\hline
Dataset / Embodiment Group & Mixture ratio \\
\hline
\endhead

\hline
\endfoot

\hline
\endlastfoot

\textbf{CMR Versius} & \textbf{4.000} \\
\quad Cholecystectomy & \\
\quad Hysterectomy & \\
\quad Inguinal hernia & \\
\quad Prostatectomy & \\

\textbf{JHU IMERSE} & \textbf{2.015} \\
\quad SRT-H porcine cholecystectomy & \\
\quad Electrocautery tumor resection & \\
\quad SutureBot & \\
\quad Needle pickup and handover & \\
\quad Suturing & \\
\quad Tissue lift & \\
\quad Pickup only & \\

\textbf{JHU ARCADE} & \textbf{0.134} \\
\quad Cholecystectomy & \\
\quad Cautery & \\

\textbf{Stanford dVRK} & \textbf{0.638} \\
\quad Needle transfer & \\
\quad Tissue retraction & \\
\quad Peg transfer & \\

\textbf{Hamlyn dVRK} & \textbf{0.396} \\
\quad Suturing-2 & \\
\quad Peg transfer & \\
\quad Suturing-1 & \\
\quad Needle grasp and handover & \\
\quad Knot tying & \\
\quad Tissue retraction & \\

\textbf{Turin MITIC} & \textbf{0.283} \\
\quad Ex vivo & \\

\textbf{UCSD dVRK} & \textbf{0.229} \\
\quad Surgical learning dataset & \\
\quad Surgical learning dataset 2 & \\

\textbf{UC Berkeley dVRK} & \textbf{0.162} \\
\quad Debridement & \\

\textbf{USTC Torin} & \textbf{0.135} \\
\quad Knot tying & \\
\quad Needle handover & \\
\quad Needle pickup & \\

\textbf{Moon Surgical} & \textbf{0.009} \\
\quad Moon & \\
\hline
\end{longtable}
\endgroup


\clearpage

\end{document}